\documentclass[10pt,twocolumn,letterpaper]{article}

\usepackage{cvpr}              

\usepackage{graphicx}
\usepackage{amsmath}
\usepackage{amssymb}
\usepackage{booktabs}

\usepackage[pagebackref,breaklinks,colorlinks]{hyperref}
\usepackage[usenames,dvipsnames]{xcolor}
\definecolor{brightgreen}{rgb}{0.4, 1.0, 0.0}
\definecolor{bostonuniversityred}{rgb}{0.8, 0.0, 0.0}
\definecolor{brickred}{rgb}{0.8, 0.25, 0.33}
\definecolor{darkspringgreen}{rgb}{0.09, 0.45, 0.27}

\newcommand{\Paragraph}[1]{\vspace{0.6mm} \noindent \textbf{#1}
\hspace{0mm}}
\newcommand{\Section}[1]{\vspace{-1mm} \section{#1} \vspace{-1mm}}
\newcommand{\SubSection}[1]{\vspace{-1mm} \subsection{#1} \vspace{-1mm}}

\usepackage[capitalize]{cleveref}
\crefname{section}{Sec.}{Secs.}
\Crefname{section}{Section}{Sections}
\Crefname{table}{Table}{Tables}
\crefname{table}{Tab.}{Tabs.}

\def\cvprPaperID{4512} 
\def\confName{CVPR}
\def\confYear{2022}

\begin{document}

\title{Face Relighting with Geometrically Consistent Shadows}

\author{Andrew Hou\thanks{All of the data mentioned in this paper was downloaded and used at Michigan State University. } $^{,1}$, Michel Sarkis$^{2}$, Ning Bi$^{2}$, Yiying Tong$^{1}$, Xiaoming Liu$^{1}$ \\
{ $^{1}$Michigan State University, $^{2}$Qualcomm Technologies Inc.
} \\
{\tt\small \{houandr1, ytong, liuxm\}@msu.edu, \{msarkis, nbi\}@qti.qualcomm.com} \\
{\small \url{https://github.com/andrewhou1/GeomConsistentFR}}
}
\maketitle

\begin{abstract}
   Most face relighting methods are able to handle diffuse shadows, but struggle to handle hard shadows, such as those cast by the nose. Methods that propose techniques for handling hard shadows often do not produce geometrically consistent shadows since they do not directly leverage the estimated face geometry while synthesizing them. We propose a novel differentiable algorithm for synthesizing hard shadows based on ray tracing, which we incorporate into training our face relighting model. Our proposed algorithm directly utilizes the estimated face geometry to synthesize geometrically consistent hard shadows. We demonstrate through quantitative and qualitative experiments on Multi-PIE and FFHQ that our method produces more geometrically consistent shadows than previous face relighting methods while also achieving state-of-the-art face relighting performance under directional lighting. In addition, we demonstrate that our differentiable hard shadow modeling improves the quality of the estimated face geometry over diffuse shading models.
\end{abstract}
\vspace{-3mm}
\Section{Introduction}
\label{sec:intro}

Single image face relighting is a problem of great interest among the computer vision and computer graphics communities. Relighting consumer photos has been a major driving factor in motivating the problem given widespread interest in photo editing. Face relighting also has applications in other areas such as Augmented Reality (AR) \cite{PhysicsGuidedRelighting}, where it can be used to modify facial illuminations to match the environment lighting, and face recognition \cite{HaLeWACV19, facerecognition}, where it can relight images to frontal illuminations. It is thus relevant both for consumer interests and entertainment and for security applications such as authentication.  

\begin{figure}[t]
\vspace{-1mm}
\begin{center}
\begin{minipage}[c]{\linewidth}
\centering
\includegraphics[width=\linewidth]{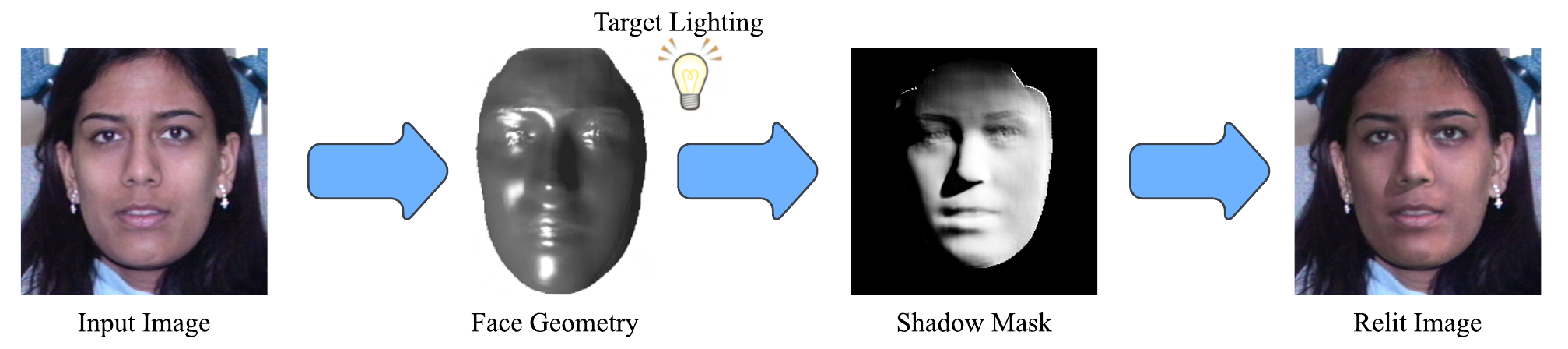} \\
\end{minipage}

\begin{minipage}[c]{0.24\linewidth}
\centering
\includegraphics[width=\linewidth]{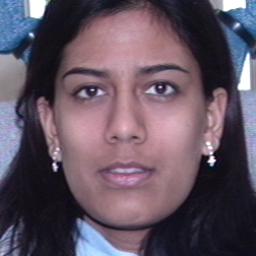} \\
\scriptsize a) Input Image
\end{minipage}
\begin{minipage}[c]{0.24\linewidth} 
\centering
\includegraphics[width=\linewidth]{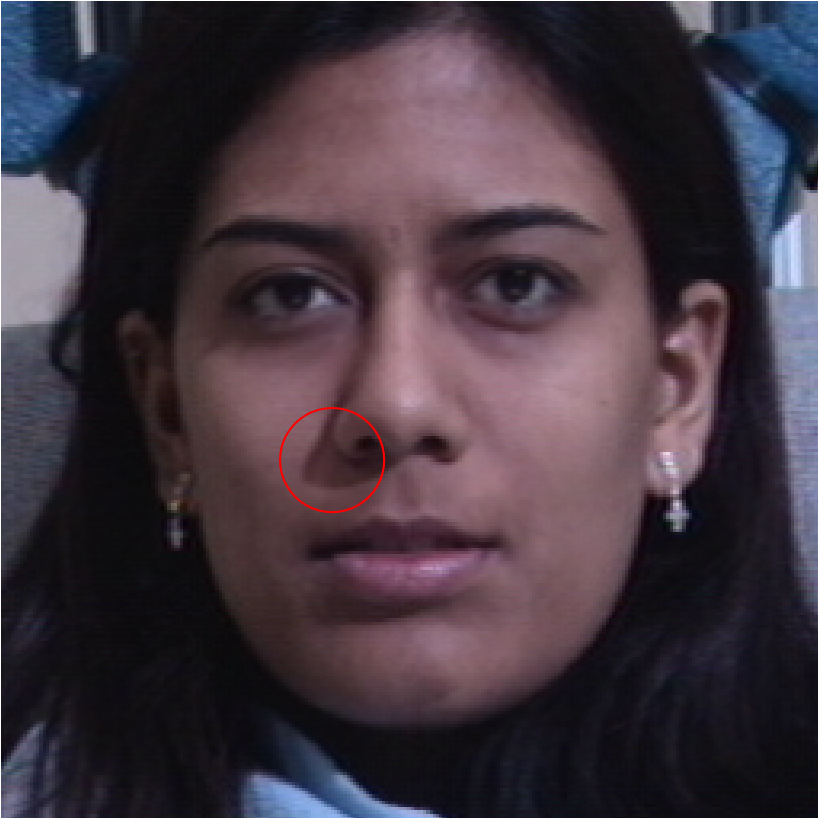} \\
\scriptsize b) Target Image
\end{minipage}
\begin{minipage}[c]{0.24\linewidth}
\centering
\includegraphics[width=\linewidth]{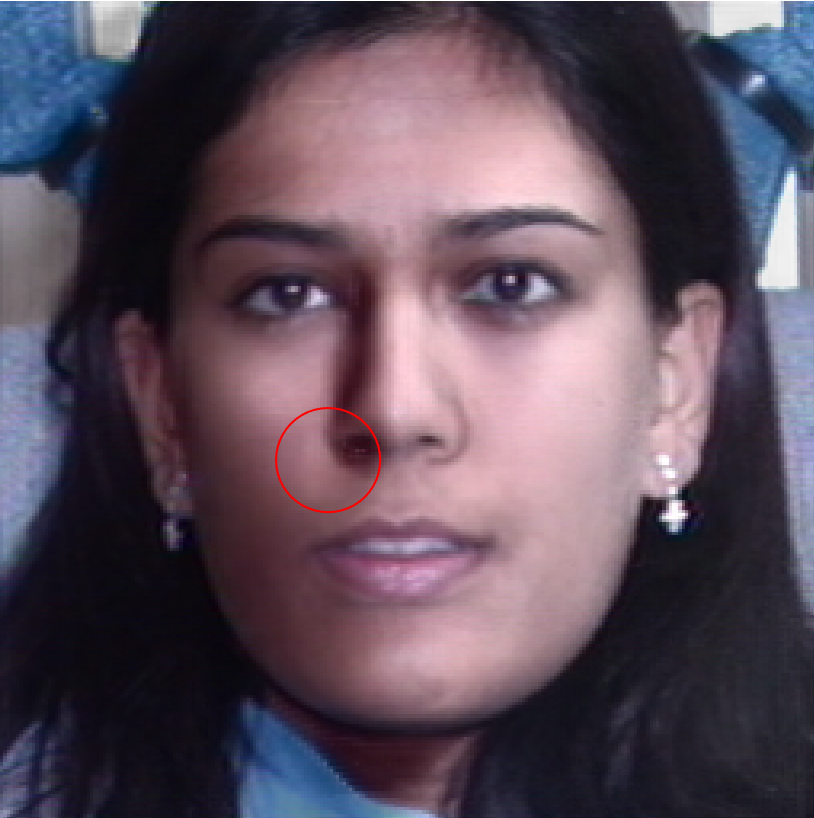} \\
\scriptsize c) Hou \cite{ShadowMaskFaceRelighting}
\end{minipage}
\begin{minipage}[c]{0.24\linewidth}
\centering
\includegraphics[width=\linewidth]{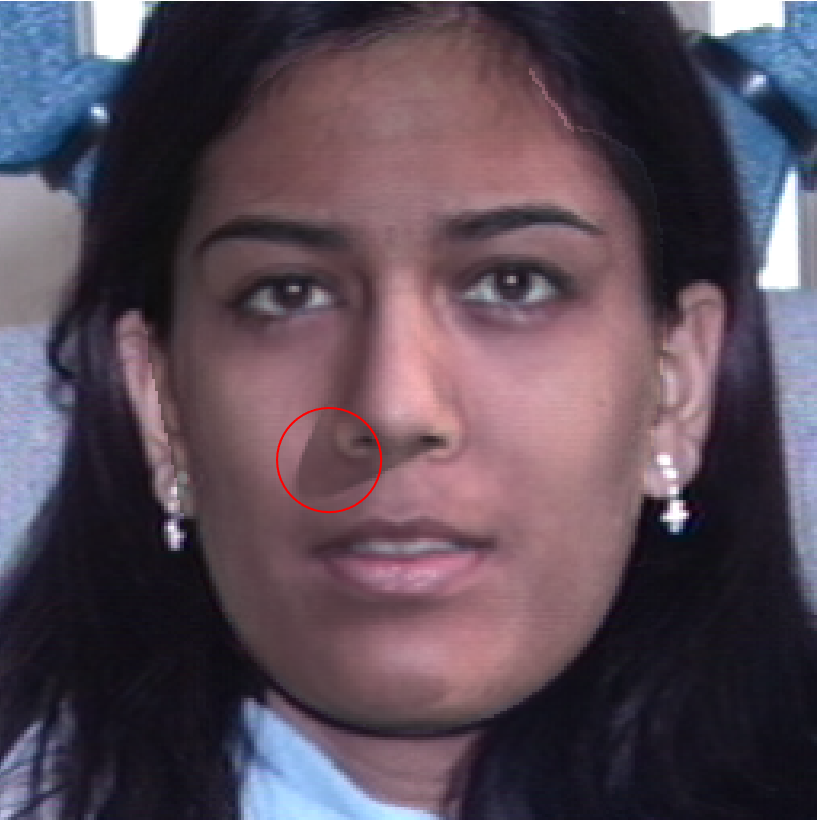} \\
\scriptsize d) Proposed
\end{minipage}
\vspace{-2mm}
\caption{\small\textbf{Overview}. We introduce a novel face relighting method that produces geometrically consistent shadows. By proposing a differentiable algorithm based on the principles of ray tracing that directly uses the face geometry for modeling hard shadows, our method produces physically correct hard shadows which the state-of-the-art face relighting method, Hou \textit{et al.} \cite{ShadowMaskFaceRelighting}, cannot produce. 
\vspace{-10mm}
}\label{fig:Fig1Overview}
\end{center}
\vspace{-3mm}
\end{figure}

Earlier relighting methods \cite{SfSNet, DPR, UCSDSingleImagePortraitRelighting} tend to make the simplified assumption that light is naturally scattered by the environment and thus diffuse in nature, and that human skin is a lambertian material. 
While this is sufficient to model general lighting directions and soft shadows, it does not account for non-lambertian effects such as hard shadows from strong directional lights. 
This is highly problematic since many light sources in the real world (\textit{e.g.} the sun) are best modeled as directional lights. 
In AR/VR, the environment lighting is often also set to be directional lights. 
In order to enhance photorealism both for in-the-wild consumer photos and in AR/VR, proper hard shadow modeling is a necessity. 

One important problem in face relighting is thus handling hard shadows. 
Most existing methods do not handle non-diffuse lighting and are unable to synthesize realistic hard shadows~\cite{DPR, UCSDSingleImagePortraitRelighting, SfSNet}. 
They generally use smooth lighting conditions such as low-order Spherical Harmonics (SH) and train on images with diffuse lighting. 
While many illumination conditions in the wild are ambient or area-based, these assumptions do not account for the interactions of strong directional lights and point lights, which produce hard shadows. 
Among current methods that do model hard shadows~\cite{ShadowMaskFaceRelighting, PhysicsGuidedRelighting, TotalRelighting}, none are able to guarantee geometrically consistent cast shadows since they do not directly utilize the estimated face geometry to generate them. 
Without using the geometry directly, the shape of the cast shadows, such as those cast by the nose, may be incorrect. 

We introduce a novel differentiable algorithm to estimate the locations of cast shadows using the principles of ray tracing. We rely on the principle that cast shadows will be located on parts of the face where the projected ray to the light source intersects some occluding surface, such as the nose. Our method can thus leverage the estimated face geometry to produce geometrically consistent hard shadows (see Fig.~\ref{fig:Fig1Overview}). We further demonstrate that differentiably modeling hard shadows can improve the quality of the face geometry, especially in regions that produce hard shadows (\textit{e.g.} the nose and near the boundary of the face), compared to models that assume diffuse shading. We therefore show that differentiable hard shadow modeling not only benefits the realism of the relit image, but also the intrinsic component estimation which can benefit other downstream tasks.

Our proposed method thus has four main contributions: 

$\diamond$ We propose a single image face relighting method that can produce geometrically consistent hard shadows. 

$\diamond$ We introduce a novel differentiable algorithm to estimate facial cast shadows based on the estimated geometry. 

$\diamond$ We achieve SoTA relighting performance on $2$ benchmarks quantitatively/qualitatively under directional lights.

$\diamond$ Our differentiable hard shadow modeling improves the estimated geometry over models that use diffuse shading. 

\Section{Related Work}
\Paragraph{Face Relighting} Prior works can be categorized into~$4$ groups:
intrinsic decomposition~\cite{MonocularNeuralReflectance, BarronMalikSfS, EggerIJCV2018, FreemanIntrinsic2018, HaLeWACV19, UncertaintyFaceReconstruction, Pfister, IntrinsicFacePriors, TowardsHighFidelityFaceReconstruction, PhysicsGuidedRelighting, SfSNet, RealisticInverseLighting, NeuralFaceEditing, color-wise-attention-network-for-low-light-image-enhancement, towards-high-fidelity-nonlinear-3d-face-morphable-model, LuanCVPR18, on-learning-3d-face-morphable-model-from-in-the-wild-images, fully-understanding-generic-objects-modeling-segmentation-and-reconstruction, WangPAMI2009, YamaguchiIntrinsic2018, TotalRelighting}, image-to-image translation~\cite{ShadowMaskFaceRelighting, UCSDSingleImagePortraitRelighting, DPR, PortraitShadowManipulation, SIPRExplicitMultipleReflectance}, style transfer~\cite{ClosedFormSolution, DeepPhotoStyleTransfer, Flickr, MassTransport, Photoapp}, and ratio images~\cite{PeersSIGGRAPH2007, ShashuaRatioImage, Stoschek2000, WenRadianceMaps}. 
Intrinsic decomposition estimates the geometry, albedo, and lighting and renders the image with a new lighting. 
Image-to-image translation instead directly estimates the relit image. 
Style transfer transfers the lighting of the reference image as a style to the input image. 
Ratio image methods relight by estimating the ratio between the source and target images or illuminations. A summary of our method compared to recent works is shown in Tab.~\ref{tab:MethodSummary}. 

Nestmeyer \textit{et al.} \cite{PhysicsGuidedRelighting} model hard shadows using a binary visibility map estimated from a U-Net. The visibility map is not directly constrained by the face geometry and therefore has a large amount of freedom, which can lead to geometrically inconsistent shadows. Also, since it is binary, their cast shadows are black whereas the true intensity of a shadow should match the environment's ambient light. 

Hou \textit{et al.} \cite{ShadowMaskFaceRelighting} utilize shadow masks to assign higher weights along hard shadow borders. While this improves the shadows, they do not use the geometry and thus the shadows can have any shape. Our model produces hard shadows where rays cast from the face to the light source intersect other parts of the face geometry, which ensures that the shadows are geometrically consistent. 

Pandey \textit{et al.} \cite{TotalRelighting} model both diffuse and specular lighting, and can generate non-lambertian effects such as hard shadows. However, they rely on a shading network to produce the relit image given the albedo and light maps, which will have network estimation error. Thus, there is no guarantee that they produce geometrically consistent shadows. 

\begin{table}
\scriptsize
\centering
\setlength\tabcolsep{2pt}
\begin{tabular}{c | c | c | c | c}
\hline
Method & \begin{tabular}{@{}c@{}}Lighting \\ Model\end{tabular} & \begin{tabular}{@{}c@{}}Model \\ Category\end{tabular} & \begin{tabular}{@{}c@{}}Handles Hard \\ Shadows\end{tabular} & \begin{tabular}{@{}c@{}}Geom.~Consistent \\ Hard Shadows\end{tabular}\\
\hline
\hline
SfSNet \cite{SfSNet} & SH & Intrinsic & \textcolor{red}{X} & \textcolor{red}{X}\\

DPR \cite{DPR} & SH & Im2Im & \textcolor{red}{X} & \textcolor{red}{X}\\

SIPR \cite{UCSDSingleImagePortraitRelighting} & \begin{tabular}{@{}c@{}}Environment \\ Map\end{tabular} & Im2Im & \textcolor{red}{X} & \textcolor{red}{X}\\

\begin{tabular}{@{}c@{}}Nestmeyer \\ \cite{PhysicsGuidedRelighting} \end{tabular} & \begin{tabular}{@{}c@{}}Directional \\ Light\end{tabular} & Intrinsic & \textcolor{green}{\checkmark} & \textcolor{red}{X}\\

Hou \cite{ShadowMaskFaceRelighting} & SH & Im2Im & \textcolor{green}{\checkmark} & \textcolor{red}{X}\\

\begin{tabular}{@{}c@{}}Total \\ Relighting~\cite{TotalRelighting} \end{tabular} & \begin{tabular}{@{}c@{}}Environment \\ Map\end{tabular} & Intrinsic & \textcolor{green}{\checkmark} & \textcolor{red}{X}\\
\hline
Proposed & \begin{tabular}{@{}c@{}}Directional \\ Light\end{tabular} & Intrinsic & \textcolor{green}{\checkmark} & \textcolor{green}{\checkmark}\\
\hline
\end{tabular}
\vspace{-2mm}
\caption{\textbf{Method Comparison}. A summary of our proposed method compared to recent face relighting methods. }
\label{tab:MethodSummary}
\vspace{-4mm}
\end{table}

\Paragraph{Differentiable Rendering and Ray Tracing}
In recent years, multiple differentiable renderers have been proposed \cite{Neural3DMeshRenderer, OpenDR, TzuMaoLi, SoftRasterizer, DifferentiableShadowComputation} that are suitable for inverse rendering tasks. However, the majority of differentiable renderers do not explicitly model shadows, particularly hard shadows. 

The most similar work to ours is Li \textit{et al.}~\cite{TzuMaoLi}, which proposes a differentiable ray tracer to model hard shadows. 
Their method operates on meshes and introduces a novel Monte Carlo edge sampling algorithm to handle the non-differentiability along triangle edges. 
Our shadow modeling is also a form of differentiable ray tracing, but operates on the $2.5$D points generated from a face depth map rather than a mesh. 
Instead of integrating over edges of mesh triangles to determine shading, we find it sufficient  to sample points between each point on the face and the light source and assign a cast shadow to the point if one of the sampled points intersects with some facial parts, such as the nose. 

Recently, Srinivasan \textit{et al.}~propose NeRV~\cite{NeRV}, which can model scene-specific hard shadows from any directions more efficiently than prior work. 
However, they rely on a visibility MLP to predict the locations of shadows, which leaves the possibility of generating geometrically inconsistent shadows due to network estimation error. 
Furthermore, NeRV requires hundreds of images with known lighting and pose and can only be trained on one static scene at a time, whereas our model is generalizable due to leveraging public face datasets in training, only requires a single image per subject, and can be applied in inference to any face image.

\Section{Proposed Method}

\begin{figure*}[t]
\begin{center}
   \includegraphics[width=0.83\linewidth]{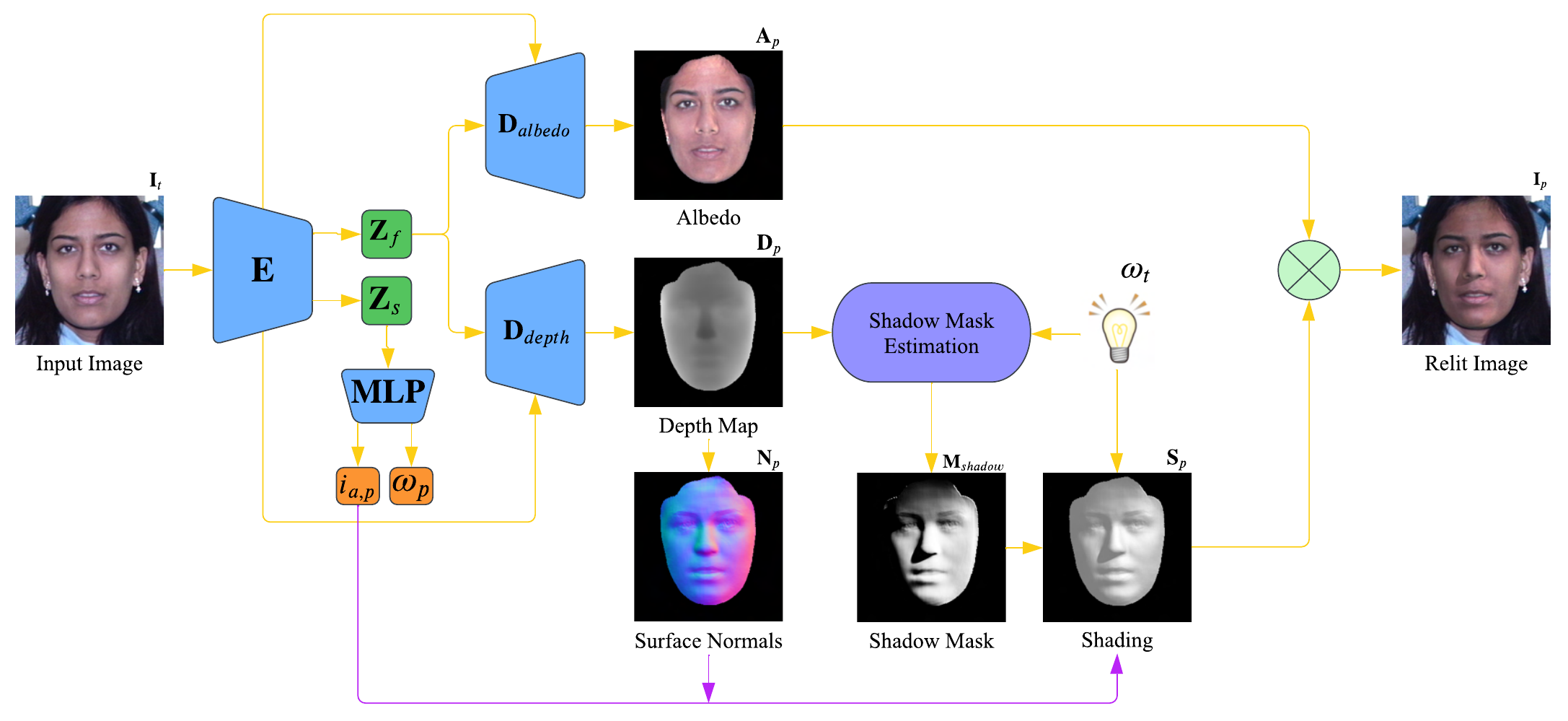} \\
   \vspace{-2mm}
\caption{\small \textbf{Model Overview}. Given a single image $\mathbf{I}_{t}$ and target lighting direction $\mathbf{\omega}_{t}$, our model generates a relit image $\mathbf{I}_{p}$ with geometrically consistent cast shadows. The geometric consistency is achieved thanks to our shadow mask estimation module, which estimates shadow mask $\mathbf{M}_{shadow}$ using depth map $\mathbf{D}_{p}$ (the face geometry). $\mathbf{M}_{shadow}$ incorporates non-diffuse cast shadows into our shading $\mathbf{S}_{p}$. 
\vspace{-7mm}}
\label{fig:Architecture}
\end{center} 
\end{figure*}

\vspace{2mm}
\SubSection{Problem Formulation}
Our relighting method relies on intrinsic decomposition, and is thus motivated by the rendering equation \cite{RenderingEquation}: 
\begin{equation}
    L_{o}(\mathbf{x}, \mathbf{\omega}_{o}) = \int_{\mathbf{\omega}_{i} \in \Omega}f(\mathbf{x}, \mathbf{\omega}_{i}, \mathbf{\omega}_{o})L_{i}(\mathbf{x}, \mathbf{\omega}_{i})\left<\mathbf{n}, \mathbf{\omega}_{i}\right>d\mathbf{\omega}_{i}, 
\end{equation}
where $\mathbf{x}$ is a point on the 3D surface, $\mathbf{n}$ is the surface normal at $\mathbf{x}$, $\mathbf{\omega}_{i}$ and $\mathbf{\omega}_{o}$ are the incoming and outgoing lighting directions respectively, $\Omega$ is the unit hemisphere centered around $\mathbf{n}$ with all possible values of $\mathbf{\omega}_{i}$, $L_{i}(\mathbf{x}, \mathbf{\omega}_{i})$ and $L_{o}(\mathbf{x}, \mathbf{\omega}_{o})$ are the incoming and outgoing radiances respectively, and $f(\mathbf{x}, \mathbf{\omega}_{i}, \mathbf{\omega}_{o})$ is the bidirectional reflectance distribution function (BRDF) determining the material's reflectance. If only diffuse reflection is considered, \textit{i.e.} $f(\mathbf{x},\mathbf{\omega}_{i},\mathbf{\omega}_{o})\!=\!a(\mathbf{x})/\pi$, the rendering equation becomes: 
\begin{equation}
    L_{o}(\mathbf{x}, \mathbf{\omega}_{o}) \!=\!
    \frac{a(\mathbf{x})}{\pi}\!\int_{\mathbf{\omega}_{i} \!\in \Omega} \! L_{i}(\mathbf{x}, \mathbf{\omega}_{i})\left<\mathbf{n}, \mathbf{\omega}_{i}\right>d\mathbf{\omega}_{i} \!=\! a(\mathbf{x})s(\mathbf{x}). 
\end{equation}
Here $a(\mathbf{x})$ is the diffuse albedo and $s(\mathbf{x})$ the diffuse shading.

Assuming there is a dominant directional light in the scene, without considering the visibility of the light source from $\mathbf{x}$, the diffuse shading $s(\mathbf{x})$ can be approximated by: 
\begin{equation}
    s(\mathbf{x}) = i_{a}+i_{d}\left<\mathbf{n}, \mathbf{\omega}_{d}\right>, 
    \label{eqn:DiffuseShading}
\end{equation}
where $i_{a}$ is the intensity of the ambient light in the scene and $i_{d}$ and $\mathbf{\omega}_{d}$ is the intensity and direction of the directional light, respectively. In other words, $L_i(\mathbf{x},\mathbf{\omega}_{i})=i_a+i_d\pi \; \delta(\mathbf{\omega}_{i},\mathbf{\omega}_{d}),$ where $\delta(\cdot,\cdot)$ is the Dirac delta function.

To model \emph{differentiable} cast shadows, we introduce the shadow mask $\mathbf{M}_{shadow}$ to model the visibility of the directional light. For each point, $\mathbf{M}_{shadow}$ stores a value close to $0$ if the point is under a cast shadow and close to $1$ otherwise. The intensity under a cast shadow should be $i_{a}$, since the directional component is blocked by some part of the face and thus only the ambient component should contribute to the shadow's intensity. To model cast shadows in the shading, we represent the modified shading $s'$ as: 
\begin{equation}
\begin{split}
    s'(\mathbf{x})&=i_{a}+\mathbf{M}_{shadow}(\mathbf{x}) i_{d}\left<\mathbf{n}, \mathbf{\omega}_{d}\right> \\ &=  \mathbf{M}_{shadow}(\mathbf{x})s(\mathbf{x})+(1-\mathbf{M}_{shadow}(\mathbf{x}))i_{a}.
\end{split} 
    \label{eqn:CastShadowShading}
\end{equation}
Our reformulation of the rendering equation is thus, 
\begin{equation}
    L_{o}(\mathbf{x}, \mathbf{\omega}_{o}) = a(\mathbf{x})s'(\mathbf{x}), 
    \label{eqn:Render}
\end{equation}
which overcomes the limitations of the diffuse shading $s$ and models cast shadows. 

Unlike prior face relighting methods that model cast shadows \cite{ShadowMaskFaceRelighting, PhysicsGuidedRelighting, TotalRelighting}, our formulation ensures geometrically consistent cast shadows by computing $\mathbf{M}_{shadow}$ directly using the estimated face geometry. We discuss this in Sec.~\ref{sec: ShadowMask}.

\SubSection{Architecture}
Given a single image $\mathbf{I}_{t}$ as input, our model estimates the intrinsic components: depth map $\mathbf{D}_{p}$ (geometry), albedo $\mathbf{A}_{p}$, the  lighting direction $\mathbf{\omega}_{p}$, and the ambient lighting intensity $i_{a, p}$. 
Our architecture is largely adopted from the hourglass network used by DPR~\cite{DPR}, but we replicate the decoder and form two branches to estimate $\mathbf{D}_{p}$ and $\mathbf{A}_{p}$ respectively. 
$\mathbf{\omega}_{p}$ and $i_{a, p}$ are estimated using a multilayer perceptron (MLP) following the encoder. 
The surface normals $\mathbf{N}_{p}$ are then computed from $\mathbf{D}_{p}$, and the shading $\mathbf{S}_{p}$ is computed via Eqn.~\ref{eqn:CastShadowShading}. 
We will discuss how $\mathbf{M}_{shadow}$ is computed in the next section. 
The final rendered image $\mathbf{I}_{p}$ is then generated following Eqn.~\ref{eqn:Render} as: 
\begin{equation}
    \mathbf{I}_{p}=\mathbf{A}_{p}\mathbf{S}_{p}. 
\end{equation}
During training, we render $\mathbf{I}_{p}$ using $\mathbf{\omega}_{p}$, the estimated lighting direction of the input image $\mathbf{I}_{t}$. We supervise the intrinsic component estimation by enforcing that $\mathbf{I}_{p}$ reproduces the input image $\mathbf{I}_{t}$. During inference, our model instead accepts a target lighting direction $\mathbf{\omega}_{t}$ as input, which allows us to perform relighting. We compute $\mathbf{M}_{shadow}$ using $\mathbf{D}_{p}$ and $\mathbf{\omega}_{t}$, which allows us to generate relit images with geometrically consistent cast shadows from any lighting direction. We illustrate our overall model architecture in Fig.~\ref{fig:Architecture}. 

\SubSection{Shading Estimation}\label{sec: ShadowMask}
One of our key contributions is producing geometrically consistent cast shadows using shadow mask $\mathbf{M}_{shadow}$. 
It is generated directly from the estimated geometry and used to produce shading $\mathbf{S}_{p}$ that models cast shadows. We now discuss the motivation and formulation. 

\Paragraph{Ray Tracing}
To motivate our algorithm for generating $\mathbf{M}_{shadow}$, we first discuss the traditional ray tracing algorithm. For every point $\mathbf{x}_{i}$ on the $3$D object, the ray tracing algorithm casts a \textit{shadow ray} towards the light source \cite{appel1968some}. If the shadow ray intersects with a surface along its path, then $\mathbf{x}_{i}$ is under cast shadow. In our setting, the shadow rays determine which points are under cast shadow based on whether the ray intersects with some parts of the estimated $3$D face, such as the nose. 

\begin{figure}[t!]
\begin{center}
   \includegraphics[width=0.9\linewidth]{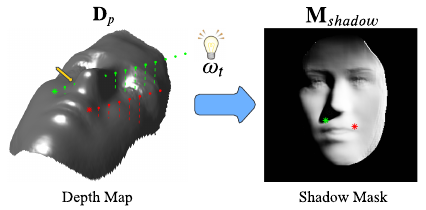} 
   \vspace{-2mm}
\caption{\small \textbf{Shadow Mask Estimation}. We generate $\mathbf{M}_{shadow}$ using $\mathbf{D}_{p}$, $\mathbf{\omega}_{t}$, and the principles of ray tracing. For every point $\mathbf{x}_{i} \in \mathbf{D}_{p}$, we sample points from $\mathbf{D}_{p}$ along the direction $\protect\overrightarrow{\mathbf{x}_{i}\mathbf{\omega}_{t}}$. If there exists a sampled point whose distance to $\protect\overrightarrow{\mathbf{x}_{i}\mathbf{\omega}_{t}}$ is close to $0$, then  $\protect\overrightarrow{\mathbf{x}_{i}\mathbf{\omega}_{t}}$ intersects a surface (\textit{e.g.} the nose) along its path and $\mathbf{x}_{i}$ is under cast shadow. If there is no such point among the sampled points, then $\mathbf{x}_{i}$ is not under cast shadow. We show $2$ points $\mathbf{x}_{1}$ and $\mathbf{x}_{2}$, marked as green and red asterisks respectively. Among the sampled points for $\mathbf{x}_{1}$ (green points), there exists a point (marked by a yellow arrow) that intersects a surface (the nose) and thus $\mathbf{x}_{1}$ is under a cast shadow. For $\mathbf{x}_{2}$, none of the sampled points (red points) intersect a surface, so $\mathbf{x}_{2}$ is not under a cast shadow. 
\vspace{-7mm}
}
\label{fig:ShadowMaskFig}
\end{center} 
\end{figure}

\Paragraph{Shadow Mask Estimation}
Our method incorporates the principles of ray tracing to generate $\mathbf{M}_{shadow}$ using the estimated depth map $\mathbf{D}_{p}$ and the target lighting direction $\mathbf{\omega}_{t}$ (see Fig.~\ref{fig:ShadowMaskFig}). 
Each pixel in $\mathbf{D}_{p}$ corresponds to a $3$D point $\mathbf{x}_{i}$, and we represent $\mathbf{\omega}_{t}$ as a unit vector in $3$D space. 
We can thus represent the shadow ray for point $\mathbf{x}_{i}$ as an order pair $(\mathbf{x}_{i},\mathbf{\omega}_{t})$. 
To determine whether it intersects the surface, we sample $m=160$ points $\mathbf{x}_{s1}, \mathbf{x}_{s2}, ..., \mathbf{x}_{sm}$ along its direction from $\mathbf{D}_{p}$ at regular intervals. 
We then provide a differentiable visibility function based on the following observation: if there exists a sampled point whose distance to the ray is close to $0$, then the current ray or a nearby ray would intersect with the surface and the point $\mathbf{x}_{i}$ is under cast shadow.
Conversely, if none of the sampled points have a distance close to $0$ to the ray, the shadow ray does not intersect the surface and $\mathbf{x}_{i}$ is not under cast shadow. 
We therefore compute the minimum distance $d_{min}$ between the sampled points and the ray by 
\begin{equation}
    d_{min} = \min \limits_{j\in[1,m]} |\overrightarrow{\mathbf{x}_{i}\mathbf{x}_{sj}}\times\mathbf{\omega}_{t}|, 
\end{equation}
where $\times$ is the cross product. 
If $d_{min}$ is close to $0$, we set the corresponding shadow mask value $\mathbf{M}_{shadow}(\mathbf{x}_{i})$ to be close to $0$, indicating $\mathbf{x}_{i}$ is under a cast shadow. Otherwise, it should be close to $1$. 
To achieve this while ensuring that computing $\mathbf{M}_{shadow}(\mathbf{x}_{i})$ is a differentiable operation, we define $\mathbf{M}_{shadow}$ as a Sigmoid function of $d_{min}$:
\begin{equation}
    \mathbf{M}_{shadow}(\mathbf{x}_{i}) = \frac{-4e^{-d_{min}}}{(1+e^{-d_{min}})^{2}}+1.
\end{equation}
We apply our algorithm to all points $\mathbf{x}_{i}$ in depth map $\mathbf{D}_{p}$ to generate the shadow mask $\mathbf{M}_{shadow}$, which indicates where cast shadows lie on the face. Since we use $\mathbf{D}_{p}$ to compute $\mathbf{M}_{shadow}$, we directly leverage the $3$D geometry of the face to synthesize our cast shadows, ensuring that they are geometrically consistent with respect to the face. 

\begin{table*}[t!]

\begin{center}
\tiny
\scalebox{1.25}{
\setlength\tabcolsep{1.5pt}{  

\begin{tabular}{ c | c | c | c | c | c | c }
\hline
Method & SfSNet \cite{SfSNet} & DPR \cite{DPR} & SIPR \cite{UCSDSingleImagePortraitRelighting} & Nestmeyer \cite{PhysicsGuidedRelighting} & Hou \cite{ShadowMaskFaceRelighting} & Proposed \\
\hline
\hline
LPIPS & $0.5222\!\pm\!0.0743$ & $0.2644\!\pm\!0.0808$ & $0.2764\!\pm\!0.0736$ & $0.3795\!\pm\!0.2294$ & $0.2013\!\pm\!0.0676$ & $\mathbf{0.1622\!\pm\!0.0490}$ \\
MSE & $0.0961\!\pm\!0.0495$ & $0.0852\!\pm\!0.0515$ & $0.0166\!\pm\!0.0107$ & $0.0588\!\pm\!0.0538$ & $0.0303\!\pm\!0.0162$ & $\mathbf{0.0150\!\pm\!0.0112}$ \\
DSSIM & $0.2918\!\pm\!0.0375$ & $0.1599\!\pm\!0.0558$ & $0.1539\!\pm\!0.0452$ & $0.2226\!\pm\!0.1356$ & $0.1186\!\pm\!0.0388$ & $\mathbf{0.0990\!\pm\!0.0381}$ \\
\hline
\end{tabular}
}}
\end{center}
\vspace{-5mm}
\caption{\small
\textbf{Relighting Evaluation on Multi-PIE Images with Target Lighting (mean$\pm$ standard deviation)}. We compare our model against methods that accept a single image and a target lighting. Our method achieves the best performance across all metrics (bold).  
}\label{tab:TargetLighting}
\vspace{2mm}
\end{table*}

\begin{table}[t!]
\begin{center}
\tiny
\scalebox{1.1}{
\setlength\tabcolsep{1.5pt}{  
\begin{tabular}{ c | c | c | c | c }
\hline
Method & Shih \cite{Flickr} & Shu \cite{MassTransport} & Hou \cite{ShadowMaskFaceRelighting} & Proposed \\
\hline
\hline
LPIPS & $0.2446\!\pm\!0.0750$ & $0.1548\!\pm\!0.0482$ & $\mathbf{0.1499\!\pm\!0.0444}$ & $0.1580\!\pm\!0.0485$ \\
MSE & $0.0529\!\pm\!0.0361$ & $0.0188\!\pm\!0.0177$ & $0.0192\!\pm\!0.0119$ & $\mathbf{0.0176\!\pm\!0.0127}$ \\
DSSIM & $0.1998 \!\pm\!0.0827$ & $0.0994\!\pm\!0.0415$ & $\mathbf{0.0942\!\pm\!0.0360}$ & $0.0962\!\pm\!0.0381$ \\
\hline
\end{tabular}
}}
\end{center}
\vspace{-5mm}
\caption{\small
\textbf{Lighting Transfer Evaluation on Multi-PIE (mean$\pm$ standard deviation)}.  Each input image is assigned a random reference image. The reference image is a different subject and a different lighting from the input image. 
\vspace{-3mm}}
\label{tab:LightingTransfer}
\end{table}

\begin{figure*}[t]
\vspace{-2mm}
\begin{center}
\begin{minipage}[c]{0.105\linewidth}
\centering
\includegraphics[width=\linewidth]{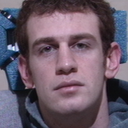}
\end{minipage}
\begin{minipage}[c]{0.105\linewidth}
\centering
\includegraphics[width=\linewidth]{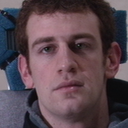}
\end{minipage}
\begin{minipage}[c]{0.105\linewidth}
\centering
\includegraphics[width=\linewidth]{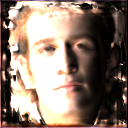}
\end{minipage}
\begin{minipage}[c]{0.105\linewidth}
\centering
\includegraphics[width=\linewidth]{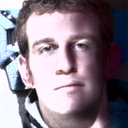}
\end{minipage}
\begin{minipage}[c]{0.105\linewidth}
\centering
\includegraphics[width=\linewidth]{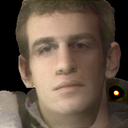}
\end{minipage}
\begin{minipage}[c]{0.105\linewidth}
\centering
\includegraphics[width=\linewidth]{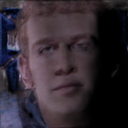}
\end{minipage}
\begin{minipage}[c]{0.105\linewidth}
\centering
\includegraphics[width=\linewidth]{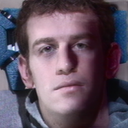}
\end{minipage}
\begin{minipage}[c]{0.105\linewidth}
\centering
\includegraphics[width=\linewidth]{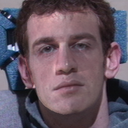}
\end{minipage}

\vspace{0.5mm}
\begin{minipage}[c]{0.105\linewidth}
\centering
\includegraphics[width=\linewidth]{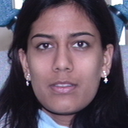} \\
\scriptsize (a) Input Image
\end{minipage}
\begin{minipage}[c]{0.105\linewidth}
\centering
\includegraphics[width=\linewidth]{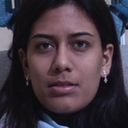} \\
\scriptsize (b) Target Image
\end{minipage}
\begin{minipage}[c]{0.105\linewidth}
\centering
\includegraphics[width=\linewidth]{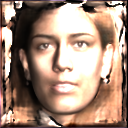}
\scriptsize (c) SfSNet \cite{SfSNet}
\end{minipage}
\begin{minipage}[c]{0.105\linewidth}
\centering
\includegraphics[width=\linewidth]{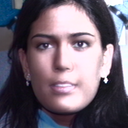}
\scriptsize (d) DPR \cite{DPR}
\end{minipage}
\begin{minipage}[c]{0.105\linewidth}
\centering
\includegraphics[width=\linewidth]{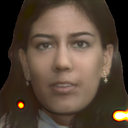} \\
\scriptsize (e) SIPR \cite{UCSDSingleImagePortraitRelighting}
\end{minipage}
\begin{minipage}[c]{0.105\linewidth}
\centering
\includegraphics[width=\linewidth]{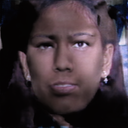} 
\scriptsize (f) Nestmeyer \cite{PhysicsGuidedRelighting}
\end{minipage}
\begin{minipage}[c]{0.105\linewidth}
\centering
\includegraphics[width=\linewidth]{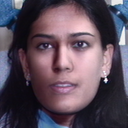} 
\scriptsize (g) Hou \cite{ShadowMaskFaceRelighting} \\
\end{minipage}
\begin{minipage}[c]{0.105\linewidth}
\centering
\includegraphics[width=\linewidth]{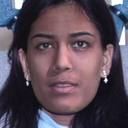} \\
\scriptsize (h) Proposed
\end{minipage}
\vspace{-2mm}
\caption{\small\textbf{Qualitative Relighting Performance on Multi-PIE (Target Lightings)}. 
Each method performs relighting given a single input image and a target lighting. 
Our method's cast shadows much more closely match the target image compared to Hou~\textit{et al.} \cite{ShadowMaskFaceRelighting} and Nestmeyer \textit{et al.}~\cite{PhysicsGuidedRelighting}, two baselines modeling cast shadows. SIPR \cite{UCSDSingleImagePortraitRelighting}, DPR \cite{DPR}, and SfSNet \cite{SfSNet} are unable to produce cast shadows. 
\vspace{-6mm}}\label{fig:MPQualitativeTargetLightings}
\end{center}

\end{figure*}

\begin{figure*}[t]
\begin{center}
\begin{minipage}[c]{0.1\linewidth}
\centering
\includegraphics[width=\linewidth]{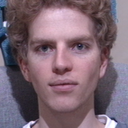}
\end{minipage}
\begin{minipage}[c]{0.1\linewidth}
\centering
\includegraphics[width=\linewidth]{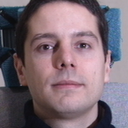}
\end{minipage}
\begin{minipage}[c]{0.1\linewidth}
\centering
\includegraphics[width=\linewidth]{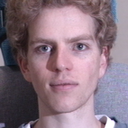}
\end{minipage}
\begin{minipage}[c]{0.1\linewidth}
\centering
\includegraphics[width=\linewidth]{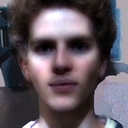}
\end{minipage}
\begin{minipage}[c]{0.1\linewidth}
\centering
\includegraphics[width=\linewidth]{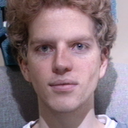}
\end{minipage}
\begin{minipage}[c]{0.1\linewidth}
\centering
\includegraphics[width=\linewidth]{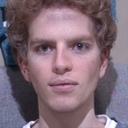} 
\end{minipage}
\begin{minipage}[c]{0.1\linewidth}
\centering
\includegraphics[width=\linewidth]{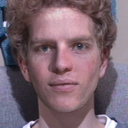}
\end{minipage}

\vspace{0.5mm}
\begin{minipage}[c]{0.1\linewidth}
\centering
\includegraphics[width=\linewidth]{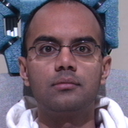} \\
\scriptsize (a) Input Image
\end{minipage}
\begin{minipage}[c]{0.1\linewidth}
\centering
\includegraphics[width=\linewidth]{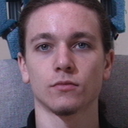} \\
\scriptsize (b) Reference
\end{minipage}
\begin{minipage}[c]{0.1\linewidth}
\centering
\includegraphics[width=\linewidth]{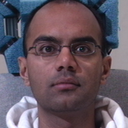} \\
\scriptsize (c) Target Image
\end{minipage}
\begin{minipage}[c]{0.1\linewidth}
\centering
\includegraphics[width=\linewidth]{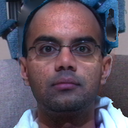} \\
\scriptsize (d) Shih \cite{Flickr}
\end{minipage}
\begin{minipage}[c]{0.1\linewidth}
\centering
\includegraphics[width=\linewidth]{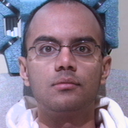} 
\scriptsize (e) Shu \cite{MassTransport}
\end{minipage}
\begin{minipage}[c]{0.1\linewidth}
\centering
\includegraphics[width=\linewidth]{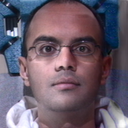} 
\scriptsize (f) Hou \cite{ShadowMaskFaceRelighting} \\
\end{minipage}
\begin{minipage}[c]{0.1\linewidth}
\centering
\includegraphics[width=\linewidth]{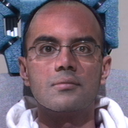} 
\scriptsize (g) Proposed \\
\end{minipage}
\vspace{-2mm}
\caption{\small\textbf{Qualitative Relighting Performance on Multi-PIE (Lighting Transfer)}. The target lighting is estimated from the reference image and used to relight the input image. Notice that our model estimates the correct target lightings from the reference images whereas Shih \textit{et al.} \cite{Flickr} transfers the wrong lightings. Furthermore, neither Shih \textit{et al.} \cite{Flickr} nor Shu \textit{et al.} \cite{MassTransport} can produce cast shadows through lighting transfer, whereas our model can. Hou \textit{et al.} \cite{ShadowMaskFaceRelighting} fails to transfer an appropriate cast shadow for the subject in the top row and the cast shadow for the subject in the second row is noticeably worse than ours in terms of shape and boundary. 
\vspace{-8mm}}\label{fig:MPQualitativeLightingTransfer}
\end{center}
\end{figure*}

\SubSection{Training Losses}
We utilize multiple loss functions to supervise the intrinsic decomposition. 
To supervise the depth estimation, we define 
    $\mathcal{L}_{depth}=\frac{\Sigma \: \mathbf{M}_{depth}\|\mathbf{D}_{p}-\mathbf{D}_{t}\|_\text{1}}{\Sigma \: \mathbf{M}_{depth}}$, 
where 
$\mathbf{D}_{t}$ is the groundtruth depth map, and mask $\mathbf{M}_{depth}$ defines where we have depth supervision in the image. 
The groundtruth depth is obtained using the method of Bai \textit{et al.} \cite{DFNRMVS} to first estimate the face mesh, and subsequently apply z-buffering. 

To supervise the albedo estimation, we define
    $\mathcal{L}_{albedo}=\frac{\Sigma \: \mathbf{M}_{face}\|\mathbf{A}_{p}-\mathbf{A}_{t}\|_\text{1}}{\Sigma \: \mathbf{M}_{face}}$, 
where 
$\mathbf{A}_{t}$ is the groundtruth albedo, and $\mathbf{M}_{face}$ is the full face mask. We generate our groundtruth albedo using SfSNet \cite{SfSNet}. Since SfSNet's estimated albedo does not generalize perfectly to our training data, we only apply $\mathcal{L}_{albedo}$ in grayscale to give our model more freedom in estimating the RGB albedo. 

To supervise our lighting estimation, we define two additional losses: $\mathcal{L}_{ambient}$ and $\mathcal{L}_{light}$. We define
    $\mathcal{L}_{ambient}=\|i_{a,p}-i_{a,t}\|_\text{1}$, 
where
$i_{a,t}$ is the groundtruth ambient intensity. 
Since determining the groundtruth ambient intensity of an image is challenging, we set $i_{a,t}$ to be the same value for all training images. We also define
    $\mathcal{L}_{light}=1-\left<\mathbf{\omega}_{p}, \: \mathbf{\omega}_{t}\right>$, 
where $\left<\mathbf{\omega}_{p}, \: \mathbf{\omega}_{t}\right>$ is the inner product between the predicted and the groundtruth lighting direction $\mathbf{\omega}_{t}$. 
We obtain $\mathbf{\omega}_{t}$ using SfSNet, and convert the estimated SH representation into a dominant lighting direction. 

To ensure that the estimated intrinsic components as a whole represent a plausible decomposition, we define a reconstruction loss $\mathcal{L}_{recon}=\frac{\Sigma \: \mathbf{M}_{face}\|\mathbf{I}_{p}-\mathbf{I}_{t}\|_\text{2}^\text{2}}{\Sigma \: \mathbf{M}_{face}}$ between the rendered image $\mathbf{I}_{p}$ and the input $\mathbf{I}_{t}$. 
Finally, to improve the perceptual quality, we employ a PatchGAN \cite{PatchGAN} discriminator that operates on $70\times70$ patches. We define our adversarial loss as $\mathcal{L}_{GAN}$ and treat our rendered images as the fake distribution and the input images as the real distribution. 
We also utilize a DSSIM loss to further improve the perceptual quality similar to Hou \textit{et al.} \cite{ShadowMaskFaceRelighting} and Nestmeyer \textit{et al.} \cite{PhysicsGuidedRelighting} defined as 
    $\mathcal{L}_{DSSIM}=\frac{(1-SSIM(\mathbf{I}_{p}, \mathbf{I}_{t}))}{2}$.
    
Our final loss function is thus defined as:
\vspace{-6mm}
\begin{center}
\begin{equation}
\begin{split}
    \mathcal{L}_{total}=\lambda_{1}\mathcal{L}_{depth}+\lambda_{2}\mathcal{L}_{albedo}+\lambda_{3}\mathcal{L}_{ambient}+ \\ \lambda_{4}\mathcal{L}_{light}+\lambda_{5}\mathcal{L}_{recon}+\lambda_{6}\mathcal{L}_{GAN}+\lambda_{7}\mathcal{L}_{DSSIM},
\end{split}
\end{equation}
\end{center}
where $\lambda_{i}$ are the weights for each loss function. 

\Paragraph{Implementation Details}\label{sec:implementationdetails}
We train using PyTorch \cite{Pytorch} for $100$ epochs with the Adam Optimizer \cite{AdamOptimizer} and a learning rate of $0.0001$ on one GeForce RTX $2080$ Ti GPU. 
In training, we set $\lambda_{1}=1$, $\lambda_{2}=5$, $\lambda_{3}=2.5$, $\lambda_{4}=1$, $\lambda_{5}=20$, $\lambda_{6}=0.01$, $\lambda_{7}=8$,
the groundtruth ambient intensity to $i_{a,t}=0.5$, and the directional intensity to $i_{d}=0.5$. 

\Section{Experiments}\label{sec:exp}
Since our training objective is to minimize the difference between the rendered image $\mathbf{I}_{p}$ and the input image $\mathbf{I}_{t}$, we are ultimately free to use any face dataset with lighting variation. 
We train our model on the CelebA-HQ dataset~\cite{CelebA-HQ}, containing $30,000$ in-the-wild face images from the CelebA dataset~\cite{CelebA}. Following the testing protocol of Hou~\textit{et al.} \cite{ShadowMaskFaceRelighting}, we evaluate our relighting performance quantitatively on the Multi-PIE~\cite{Multi-PIE} dataset, which contains $18$ images per subject each with a unique directional light. 

\SubSection{Quantitative Evaluations}
\Paragraph{Multi-PIE Evaluation with Target Lightings} As our model relights using a target lighting $\mathbf{\omega}_{t}$, we randomly select $1$ of the $18$ directional lights in Multi-PIE as $\mathbf{\omega}_{t}$ for each subject and also randomly select $1$ of the $18$ images as the input image. 
Since Multi-PIE captures each subject under every lighting condition, we have the relighting groundtruth and can quantitatively compare our relit image with the groundtruth image under the target lighting $\mathbf{\omega}_{t}$. 
We thus compare with prior methods that accept a target lighting as input \cite{SfSNet, PhysicsGuidedRelighting, ShadowMaskFaceRelighting, DPR, UCSDSingleImagePortraitRelighting}. 
We evaluate using $3$ metrics: MSE, DSSIM \cite{PhysicsGuidedRelighting}, and LPIPS \cite{LPIPS}. 
Both DSSIM and LPIPS are metrics that are highly correlated with perceptual quality~\cite{LPIPS, PhysicsGuidedRelighting}. 
$\text{DSSIM} = \frac{1}{2}(1-\text{SSIM})$ is an error metric defined based on SSIM~\cite{SSIM}. 
During evaluation, we compute the metrics for all methods only in the face region indicated by $\mathbf{M}_{depth}$. 
This ensures a fair comparison with our method, since $\mathbf{M}_{depth}$ represents where our images receive depth supervision from Bai~\textit{et al.} \cite{DFNRMVS}, which does not estimate the depth outside of the face. 
Our method is thus intended to relight the face region, not the hair or background. 

We report the results of our evaluation in Tab.~\ref{tab:TargetLighting} and note that our model achieves the best performance across all $3$ metrics, with the largest gains in DSSIM and LPIPS, indicating that the perceptual quality of our relit images significantly improves over prior work. 
This is largely due to our differentiable hard shadow modeling that generates more appropriately shaped hard shadows. We also explicitly model both ambient and directional light which helps to produce more well-balanced colors in our relit images than prior work, where the shadows may be too dark or the illuminated face may be too bright.   

\Paragraph{Multi-PIE Evaluation Using Lighting Transfer} 
Some methods require both an input and a reference image and relight by transferring the style of the reference to the input, known as lighting transfer. 
To evaluate our lighting transfer performance, we sample a random lighting for each Multi-PIE subject to serve as the input and a random reference image from the entire dataset. 
The reference image is a different subject from the input and under a different lighting. 
For lighting transfer, we first feed the reference image to our model to estimate the target lighting direction $\mathbf{\omega}_{t}$ and the ambient intensity $i_{a, p}$. 
We then pass the input image along with $\mathbf{\omega}_{t}$ and $i_{a, p}$ to our model to generate the relit image. 
The groundtruth target image is readily available in Multi-PIE.
We compare with Shih \textit{et al.} \cite{Flickr}, Shu \textit{et al.} \cite{MassTransport}, and Hou \textit{et al.} \cite{ShadowMaskFaceRelighting} and report the results in Tab.~\ref{tab:LightingTransfer}. We achieve the best performance in MSE and comparable performance to Hou \textit{et al.} in terms of DSSIM and LPIPS. We believe that a large reason why the performance is lower is the imperfect lighting supervision from SfSNet \cite{SfSNet}, which limits our model's ability to estimate the correct lighting from the reference image and lowers the relighting performance. 

\SubSection{Qualitative Evaluations}
\Paragraph{Multi-PIE Results} We show qualitative relighting results of Multi-PIE \cite{Multi-PIE} subjects for target lightings in Fig.~\ref{fig:MPQualitativeTargetLightings} and for lighting transfer in  Fig.~\ref{fig:MPQualitativeLightingTransfer}. 
When using target lightings, our model produces cast shadows that much more closely match the shape of the shadows in target images compared to prior work. 
This is due to our shadow mask estimation that incorporates the face geometry to synthesize cast shadows, improving their geometric consistency. 
Hou \textit{et al.} \cite{ShadowMaskFaceRelighting} and Nestmeyer \textit{et al.} \cite{PhysicsGuidedRelighting} also model cast shadows, but neither synthesize cast shadows directly from the face geometry and instead regress them from CNNs. 
Thus, they have no guarantee that their cast shadows correctly match the face geometry, as seen in Fig.~\ref{fig:MPQualitativeTargetLightings}.
SIPR~\cite{UCSDSingleImagePortraitRelighting}, DPR \cite{DPR}, and SfSNet \cite{SfSNet} primarily model diffuse lightings and generally cannot produce cast shadows. 
When performing lighting transfer, we notice that our model can accurately estimate the target lighting from the reference image and is superior to the baselines in transferring over cast shadows (see Fig.~\ref{fig:MPQualitativeLightingTransfer}). 
Shih \textit{et al.} \cite{Flickr} and Shu \textit{et al.} \cite{MassTransport} largely fail to transfer over cast shadows and Hou \textit{et al.} \cite{ShadowMaskFaceRelighting} produces cast shadows that do not match the shape of the groundtruth. 
\vspace{6mm}

\begin{figure}[t!]
\vspace{-1mm}
\begin{center}
   \includegraphics[width=0.85\linewidth]{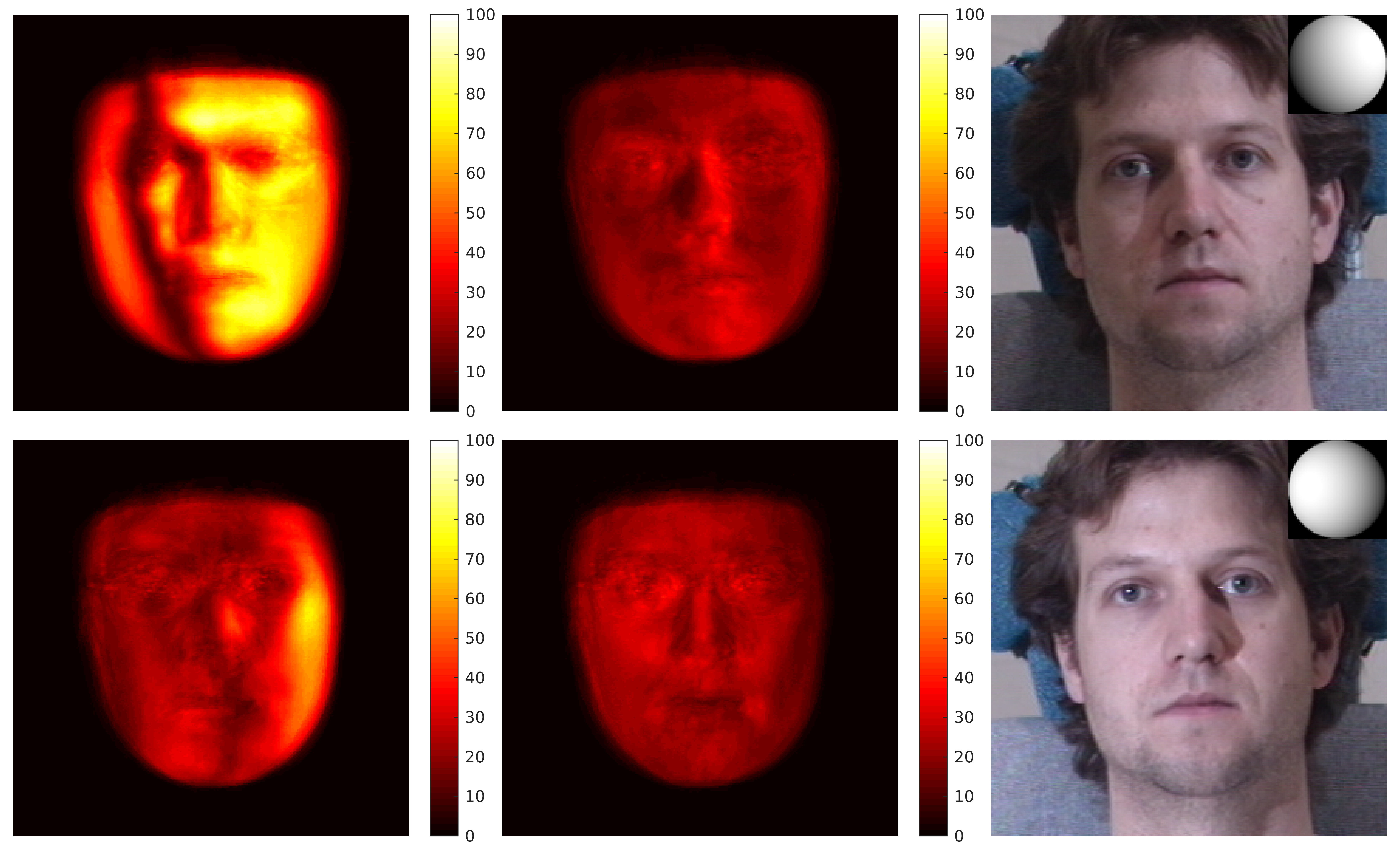} \\
   \scriptsize \quad\:\:\: a) Hou \cite{ShadowMaskFaceRelighting} \quad\quad\quad\quad\:\:\: b) Proposed \quad\quad\quad\:\:\: c) Target Lighting
   \vspace{-2mm}
\caption{\small \textbf{Relighting Error Maps}. 
We show the average $L_{1}$ error map between our relit images and the groundtruth test images of Multi-PIE for each lighting and compare with Hou \textit{et al.} \cite{ShadowMaskFaceRelighting}. 
As shown in b), we have lower error in the shadowed regions, including shadows cast around the nose. 
Hou \textit{et al.}~has higher errors around the cast shadows, demonstrating that our method produces more geometrically consistent shadows across all subjects. \vspace{-8mm}}
\label{fig:ErrorMap}
\end{center} 
\end{figure}
\vspace{-5mm}

\begin{figure*}[t]
\vspace{-2mm}
\begin{center}
\begin{minipage}[c]{0.107\linewidth}
\centering
\includegraphics[width=\linewidth]{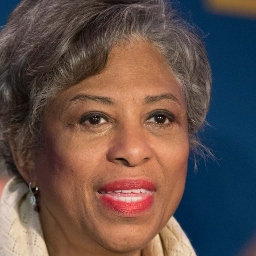}
\end{minipage}
\begin{minipage}[c]{0.107\linewidth}
\centering
\includegraphics[width=\linewidth]{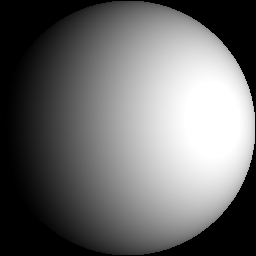}
\end{minipage}
\begin{minipage}[c]{0.107\linewidth}
\centering
\includegraphics[width=\linewidth]{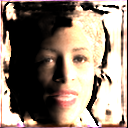}
\end{minipage}
\begin{minipage}[c]{0.107\linewidth}
\centering
\includegraphics[width=\linewidth]{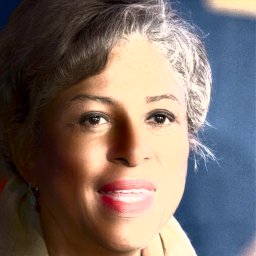}
\end{minipage}
\begin{minipage}[c]{0.107\linewidth}
\centering
\includegraphics[width=\linewidth]{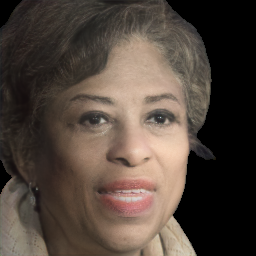}
\end{minipage}
\begin{minipage}[c]{0.107\linewidth}
\centering
\includegraphics[width=\linewidth]{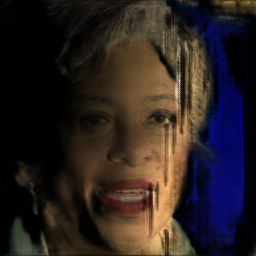}
\end{minipage}
\begin{minipage}[c]{0.107\linewidth}
\centering
\includegraphics[width=\linewidth]{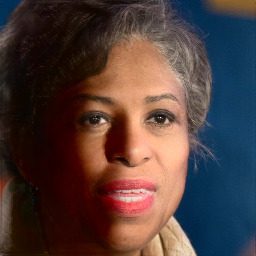}
\end{minipage}
\begin{minipage}[c]{0.107\linewidth}
\centering
\includegraphics[width=\linewidth]{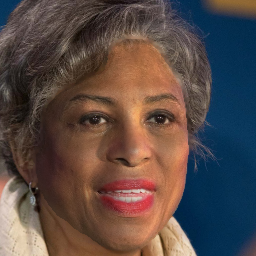}
\end{minipage}

\begin{minipage}[c]{0.107\linewidth}
\centering
\includegraphics[width=\linewidth]{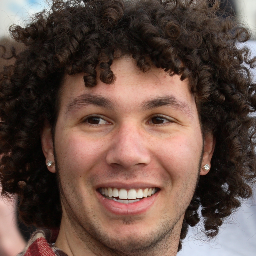}
\end{minipage}
\begin{minipage}[c]{0.107\linewidth}
\centering
\includegraphics[width=\linewidth]{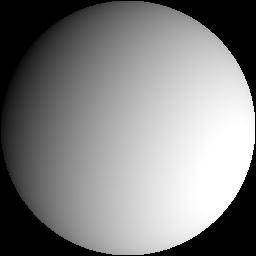}
\end{minipage}
\begin{minipage}[c]{0.107\linewidth}
\centering
\includegraphics[width=\linewidth]{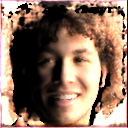}
\end{minipage}
\begin{minipage}[c]{0.107\linewidth}
\centering
\includegraphics[width=\linewidth]{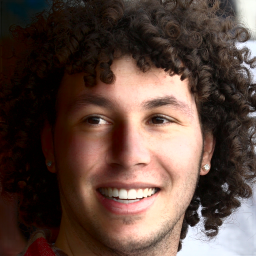}
\end{minipage}
\begin{minipage}[c]{0.107\linewidth}
\centering
\includegraphics[width=\linewidth]{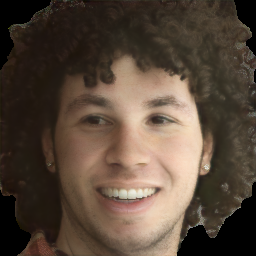}
\end{minipage}
\begin{minipage}[c]{0.107\linewidth}
\centering
\includegraphics[width=\linewidth]{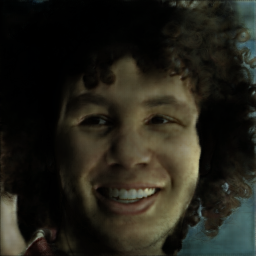}
\end{minipage}
\begin{minipage}[c]{0.107\linewidth}
\centering
\includegraphics[width=\linewidth]{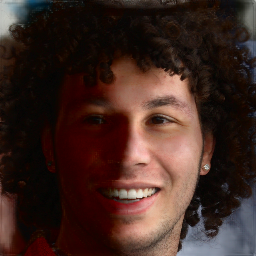}
\end{minipage}
\begin{minipage}[c]{0.107\linewidth}
\centering
\includegraphics[width=\linewidth]{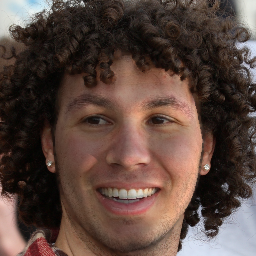}
\end{minipage}

\vspace{0.5mm}
\begin{minipage}[c]{0.107\linewidth}
\centering
\includegraphics[width=\linewidth]{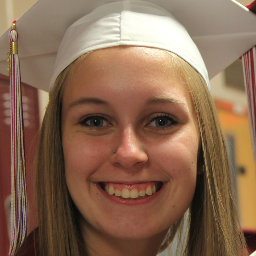}
\end{minipage}
\begin{minipage}[c]{0.107\linewidth}
\centering
\includegraphics[width=\linewidth]{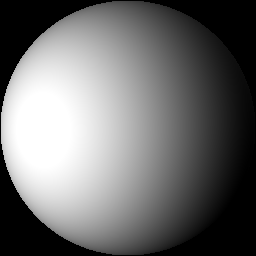}
\end{minipage}
\begin{minipage}[c]{0.107\linewidth}
\centering
\includegraphics[width=\linewidth]{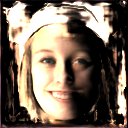}
\end{minipage}
\begin{minipage}[c]{0.107\linewidth}
\centering
\includegraphics[width=\linewidth]{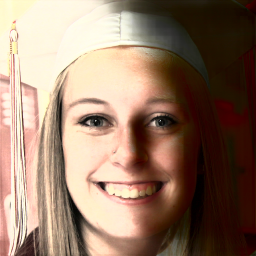}
\end{minipage}
\begin{minipage}[c]{0.107\linewidth}
\centering
\includegraphics[width=\linewidth]{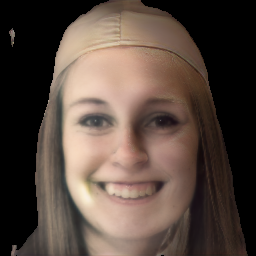}
\end{minipage}
\begin{minipage}[c]{0.107\linewidth}
\centering
\includegraphics[width=\linewidth]{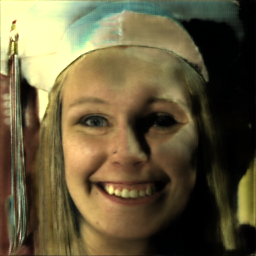}
\end{minipage}
\begin{minipage}[c]{0.107\linewidth}
\centering
\includegraphics[width=\linewidth]{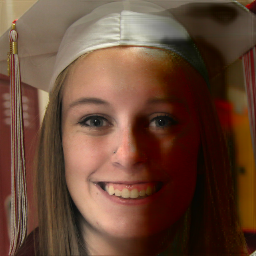}
\end{minipage}
\begin{minipage}[c]{0.107\linewidth}
\centering
\includegraphics[width=\linewidth]{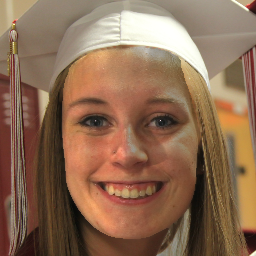}
\end{minipage}

\vspace{0.5mm}
\begin{minipage}[c]{0.107\linewidth}
\centering
\includegraphics[width=\linewidth]{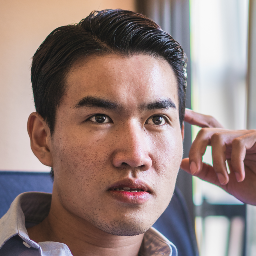}
\scriptsize (a) Input Image
\end{minipage}
\begin{minipage}[c]{0.107\linewidth}
\centering
\includegraphics[width=\linewidth]{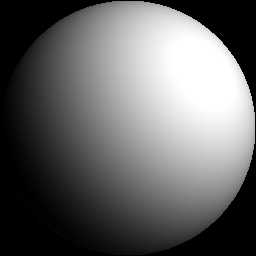}
\scriptsize (b) Target Lighting
\end{minipage}
\begin{minipage}[c]{0.107\linewidth}
\centering
\includegraphics[width=\linewidth]{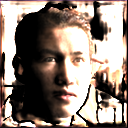}
\scriptsize (c) SfSNet \cite{SfSNet}
\end{minipage}
\begin{minipage}[c]{0.107\linewidth}
\centering
\includegraphics[width=\linewidth]{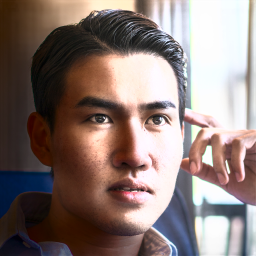}
\scriptsize (d) DPR \cite{DPR}
\end{minipage}
\begin{minipage}[c]{0.107\linewidth}
\centering
\includegraphics[width=\linewidth]{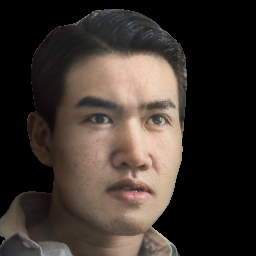}
\scriptsize (e) SIPR \cite{UCSDSingleImagePortraitRelighting}
\end{minipage}
\begin{minipage}[c]{0.107\linewidth}
\centering
\includegraphics[width=\linewidth]{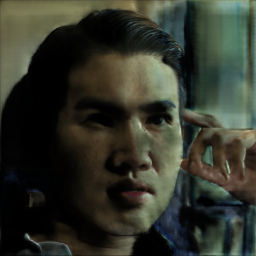}
\scriptsize (f) Nestmeyer \cite{PhysicsGuidedRelighting}
\end{minipage}
\begin{minipage}[c]{0.107\linewidth}
\centering
\includegraphics[width=\linewidth]{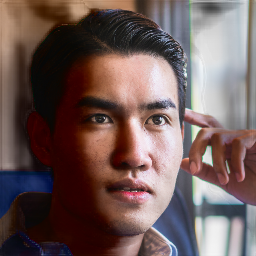}
\scriptsize (g) Hou \cite{ShadowMaskFaceRelighting}
\end{minipage}
\begin{minipage}[c]{0.107\linewidth}
\centering
\includegraphics[width=\linewidth]{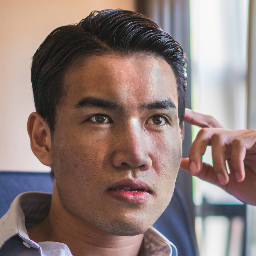}
\scriptsize (h) Proposed
\end{minipage}

\vspace{-2mm}
\caption{\small\textbf{Qualitative Relighting Performance on FFHQ}. Across multiple in-the-wild subjects and target lightings, our model produces more geometrically consistent cast shadows than prior methods while achieving noticeably better visual quality. Best viewed if enlarged. 
\vspace{-4mm}}\label{fig:FFHQQualitativeMultipleSubjects}
\end{center}

\end{figure*}

\begin{figure*}[t]
\begin{center}
\begin{minipage}[c]{0.02\linewidth}
a)
\end{minipage}
\begin{minipage}[c]{0.095\linewidth}
\centering
\includegraphics[width=\linewidth]{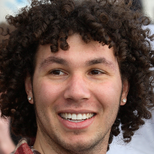}
\end{minipage}
\begin{minipage}[c]{0.095\linewidth}
\centering
\includegraphics[width=\linewidth]{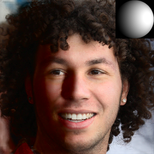}
\end{minipage}
\begin{minipage}[c]{0.095\linewidth}
\centering
\includegraphics[width=\linewidth]{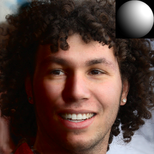}
\end{minipage}
\begin{minipage}[c]{0.095\linewidth}
\centering
\includegraphics[width=\linewidth]{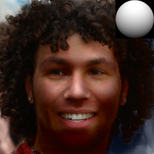}
\end{minipage}
\begin{minipage}[c]{0.095\linewidth}
\centering
\includegraphics[width=\linewidth]{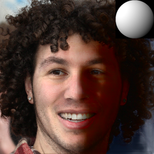}
\end{minipage}
\begin{minipage}[c]{0.095\linewidth}
\centering
\includegraphics[width=\linewidth]{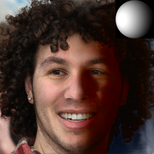}
\end{minipage}
\begin{minipage}[c]{0.095\linewidth}
\centering
\includegraphics[width=\linewidth]{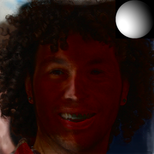}
\end{minipage}
\begin{minipage}[c]{0.095\linewidth}
\centering
\includegraphics[width=\linewidth]{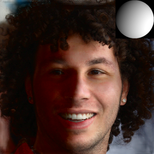}
\end{minipage}

\vspace{0.5mm}
\begin{minipage}[c]{0.02\linewidth}
b)
\end{minipage}
\begin{minipage}[c]{0.095\linewidth}
\centering
\includegraphics[width=\linewidth]{Figures/Qualitative_FFHQ_Single_Subject/00508_input_compressed.png}
\scriptsize Input Image
\end{minipage}
\begin{minipage}[c]{0.095\linewidth}
\centering
\includegraphics[width=\linewidth]{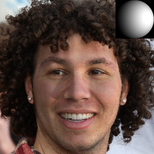}
\scriptsize Light $1$
\end{minipage}
\begin{minipage}[c]{0.095\linewidth}
\centering
\includegraphics[width=\linewidth]{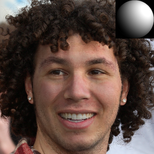}
\scriptsize Light $2$
\end{minipage}
\begin{minipage}[c]{0.095\linewidth}
\centering
\includegraphics[width=\linewidth]{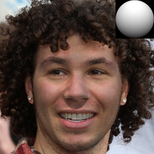}
\scriptsize Light $3$
\end{minipage}
\begin{minipage}[c]{0.095\linewidth}
\centering
\includegraphics[width=\linewidth]{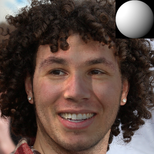}
\scriptsize Light $4$
\end{minipage}
\begin{minipage}[c]{0.095\linewidth}
\centering
\includegraphics[width=\linewidth]{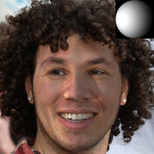}
\scriptsize Light $5$
\end{minipage}
\begin{minipage}[c]{0.095\linewidth}
\centering
\includegraphics[width=\linewidth]{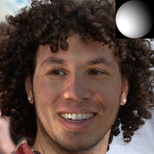}
\scriptsize Light $6$
\end{minipage}
\begin{minipage}[c]{0.095\linewidth}
\centering
\includegraphics[width=\linewidth]{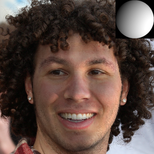}
\scriptsize Light $7$
\end{minipage}
\vspace{-2mm}
\caption{\small\textbf{Comparison of Geometric Consistency of Cast Shadows}. 
We compare the geometric consistency of our cast shadows across $7$ target lightings with Hou \textit{et al.} \cite{ShadowMaskFaceRelighting}, another face relighting method that models cast shadows. Notice that our model's cast shadows shown in row b) are more plausible in terms of shape and shadow boundaries than the cast shadows of Hou \textit{et al.}, shown in row a). 
\vspace{-7mm}}\label{fig:FFHQQualitativeSingleSubject}
\end{center}
\end{figure*}

\Paragraph{FFHQ Results} We evaluate our performance on in-the-wild faces from the FFHQ~\cite{FFHQ} dataset. 
We show in Fig.~\ref{fig:FFHQQualitativeMultipleSubjects} that we produce more geometrically consistent shadows than prior work across several subjects and lightings. 
We also show in Fig.~\ref{fig:FFHQQualitativeSingleSubject} that our  cast shadows are geometrically consistent as we rotate the target lightings around the face. 
Compared to Hou \textit{et al.} \cite{ShadowMaskFaceRelighting}, our cast shadows have much more plausible shapes and shadow boundaries.
\vspace{2mm}

\begin{figure*}[t]
\begin{center}
\begin{minipage}[c]{0.09\linewidth}
\centering
\includegraphics[width=\linewidth]{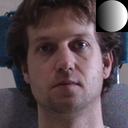}
\tiny Light $1$ \textcolor{darkspringgreen}{($7.80$\%)}\\
\tiny P: $\mathbf{11.6537}$ \\
\tiny D: $12.6395$ \\
\tiny S: $14.9658$\\
\end{minipage}
\begin{minipage}[c]{0.09\linewidth}
\centering
\includegraphics[width=\linewidth]{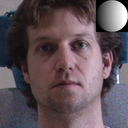}
\tiny Light $2$ \textcolor{brightgreen}{($12.60$\%)}\\
\tiny P: $\mathbf{10.9290}$ \\
\tiny D: $12.5049$ \\
\tiny S: $13.7211$\\
\end{minipage}
\begin{minipage}[c]{0.09\linewidth}
\centering
\includegraphics[width=\linewidth]{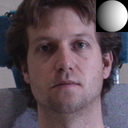}
\tiny Light $3$ \textcolor{brightgreen}{($10.60$\%)}\\
\tiny P: $\mathbf{11.5282}$ \\
\tiny D: $12.8954$ \\
\tiny S: $14.6772$\\
\end{minipage}
\begin{minipage}[c]{0.09\linewidth}
\centering
\includegraphics[width=\linewidth]{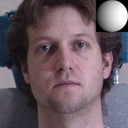}
\tiny Light $4$ \textcolor{brightgreen}{($10.25$\%)}\\
\tiny P: $\mathbf{10.5350}$ \\
\tiny D: $11.7384$ \\
\tiny S: $13.9137$\\
\end{minipage}
\begin{minipage}[c]{0.09\linewidth}
\centering
\includegraphics[width=\linewidth]{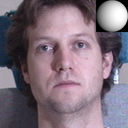}
\tiny Light $5$ \textcolor{darkspringgreen}{($8.72$\%)}\\
\tiny P: $\mathbf{10.8613}$ \\
\tiny D: $11.8990$ \\
\tiny S: $13.8324$\\
\end{minipage}
\begin{minipage}[c]{0.09\linewidth}
\centering
\includegraphics[width=\linewidth]{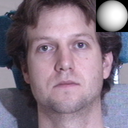}
\tiny Light $6$ \textcolor{brightgreen}{($14.50$\%)}\\
\tiny P: $\mathbf{13.0104}$ \\
\tiny D: $15.2162$ \\
\tiny S: $15.9875$\\
\end{minipage}
\begin{minipage}[c]{0.09\linewidth}
\centering
\includegraphics[width=\linewidth]{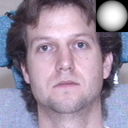}
\tiny Light $7$ \textcolor{darkspringgreen}{($4.44$\%)}\\
\tiny P: $\mathbf{10.3723}$ \\
\tiny D: $10.8543$ \\
\tiny S: $13.3527$\\
\end{minipage}
\begin{minipage}[c]{0.09\linewidth}
\centering
\includegraphics[width=\linewidth]{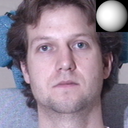}
\tiny Light $8$ \textcolor{brightgreen}{($16.10$\%)}\\
\tiny P: $\mathbf{10.8003}$ \\
\tiny D: $12.9457$ \\
\tiny S: $12.8724$\\
\end{minipage}
\begin{minipage}[c]{0.09\linewidth}
\centering
\includegraphics[width=\linewidth]{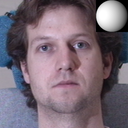}
\tiny Light $9$ \textcolor{brightgreen}{($10.54$\%)}\\
\tiny P: $\mathbf{10.3903}$ \\
\tiny D: $11.6144$ \\
\tiny S: $13.8066$\\
\end{minipage}

\begin{minipage}[c]{0.09\linewidth}
\centering
\includegraphics[width=\linewidth]{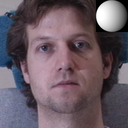}
\tiny Light $10$ \textcolor{brightgreen}{($14.69$\%)}\\
\tiny P: $\mathbf{12.1590}$ \\
\tiny D: $14.2524$ \\
\tiny S: $15.4159$\\
\end{minipage}
\begin{minipage}[c]{0.09\linewidth}
\centering
\includegraphics[width=\linewidth]{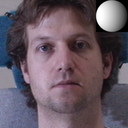}
\tiny Light $11$ \textcolor{darkspringgreen}{($6.63$\%)}\\
\tiny P: $\mathbf{12.2808}$ \\
\tiny D: $13.1525$ \\
\tiny S: $15.5531$\\
\end{minipage}
\begin{minipage}[c]{0.09\linewidth}
\centering
\includegraphics[width=\linewidth]{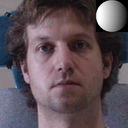}
\tiny Light $12$ \textcolor{darkspringgreen}{($8.68$\%)}\\
\tiny P: $\mathbf{10.4933}$ \\
\tiny D: $11.4909$ \\
\tiny S: $15.2955$\\
\end{minipage}
\begin{minipage}[c]{0.09\linewidth}
\centering
\includegraphics[width=\linewidth]{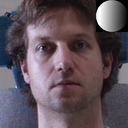}
\tiny Light $13$ \textcolor{darkspringgreen}{($8.48$\%)}\\
\tiny P: $\mathbf{11.6369}$ \\
\tiny D: $12.7155$ \\
\tiny S: $14.4261$\\
\end{minipage}
\begin{minipage}[c]{0.09\linewidth}
\centering
\includegraphics[width=\linewidth]{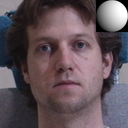}
\tiny Light $14$ \textcolor{darkspringgreen}{($5.94$\%)}\\
\tiny P: $\mathbf{11.3081}$ \\
\tiny D: $12.0220$ \\
\tiny S: $14.6807$\\
\end{minipage}
\begin{minipage}[c]{0.09\linewidth}
\centering
\includegraphics[width=\linewidth]{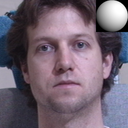}
\tiny Light $15$ \textcolor{brightgreen}{($13.76$\%)}\\
\tiny P: $\mathbf{10.2569}$ \\
\tiny D: $11.8937$ \\
\tiny S: $12.7515$\\
\end{minipage}
\begin{minipage}[c]{0.09\linewidth}
\centering
\includegraphics[width=\linewidth]{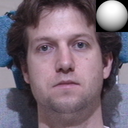}
\tiny Light $16$ \textcolor{darkspringgreen}{($8.99$\%)}\\
\tiny P: $\mathbf{10.8084}$ \\
\tiny D: $11.8756$ \\
\tiny S: $13.8700$\\
\end{minipage}
\begin{minipage}[c]{0.09\linewidth}
\centering
\includegraphics[width=\linewidth]{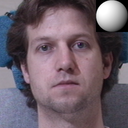}
\tiny Light $17$ \textcolor{brightgreen}{($14.93$\%)}\\
\tiny P: $\mathbf{11.0435}$ \\
\tiny D: $12.9818$ \\
\tiny S: $13.8895$\\
\end{minipage}
\begin{minipage}[c]{0.09\linewidth}
\centering
\includegraphics[width=\linewidth]{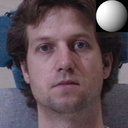}
\tiny Light $18$ \textcolor{brightgreen}{($12.55$\%)}\\
\tiny P: $\mathbf{10.4406}$ \\
\tiny D: $11.9396$ \\
\tiny S: $14.4629$\\
\end{minipage}
\vspace{-2mm}
\caption{\small\textbf{Per Lighting Surface Normal Error on Multi-PIE}. 
We compute the surface normal errors (degrees) across all test images of the same lighting. 
Our proposed model (P) achieves lower errors across all lightings than DFNRMVS \cite{DFNRMVS} (D)  
and SfSNet \cite{SfSNet} (S).  
We record the  improvement percentage of P over the best baseline, where higher percentages are highlighted in brighter green. 
A reference image and its lighting direction are provided for each lighting. 
P improves the normals the most for lightings with hard shadows (\textit{e.g.} lightings $4$, $6$, $8$, $9$, $10$, $15$, $17$, and $18$), which highlights the benefit of our hard shadow modeling in improving the face geometry.   
\vspace{-7mm} }\label{fig:GeometryAbsoluteDepth}
\end{center}
\vspace{1mm}
\end{figure*}

\SubSection{Ablations and Additional Experiments}
\Paragraph{Reconstruction Error Analysis} To better understand the distribution of reconstruction errors during relighting, Fig.~\ref{fig:ErrorMap} visualizes the average $L_{1}$ error map between our relit images and the groundtruth target images in our Multi-PIE test set. 
We generate an error map for each target lighting separately and compute the average across all test subjects with that target lighting. 
We compare our error maps with Hou \textit{et al.} \cite{ShadowMaskFaceRelighting}, the SoTA face relighting method, and notice that our error maps have much lower error in the shadowed face regions, including the shadows cast around the nose. 
This further demonstrates that our method produces more geometrically consistent shadows across all test subjects.

\Paragraph{Geometry Error Analysis} 
One benefit of modeling hard shadows differentiably is that the end-to-end training may improve the intrinsic components, such as geometry, in face regions that cast hard shadows. 
To demonstrate this, we compare our surface normal errors on the Multi-PIE test images with two baselines: SfSNet~\cite{SfSNet}, an intrinsic decomposition method with a diffuse SH lighting model, and DFNRMVS~\cite{DFNRMVS}, which provides our geometry supervision. 
We choose surface normal error as the metric since the rendering equation uses surface normals, rather than the depth, to compute the shading. 
Although Multi-PIE lacks groundtruth $3$D shapes, we use DFNRMVS~\cite{DFNRMVS} to estimate face meshes given $3$ multi-view faces per subject as input, from which we compute the groundtruth surface normals. 
A dataset with large lighting variation and $3$D groundtruth shapes is still lacking, partially due to the sensitivity of $3$D scanners to illumination. 
 We train on Multi-PIE subjects $1$-$250$ and test on subjects $251$-$346$. 
 As for the geometry supervision in training, we use the face meshes from DFNRMVS provided only a single frontal image as input, which produces lower quality shapes than $3$ views. 
 
\begin{figure}[t]
\vspace{-1mm}
\begin{center}
\begin{minipage}[c]{0.2\linewidth}
\centering
\includegraphics[width=\linewidth]{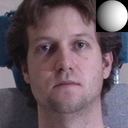}
\end{minipage}
\begin{minipage}[c]{0.2\linewidth}
\centering
\includegraphics[width=\linewidth]{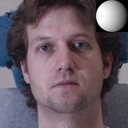}
\end{minipage}
\begin{minipage}[c]{0.2\linewidth}
\centering
\includegraphics[width=\linewidth]{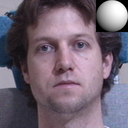}
\end{minipage}
\begin{minipage}[c]{0.2\linewidth}
\centering
\includegraphics[width=\linewidth]{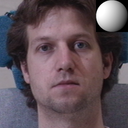}
\end{minipage}

\vspace{0.5mm}
\begin{minipage}[c]{0.2\linewidth}
\centering
\includegraphics[width=\linewidth]{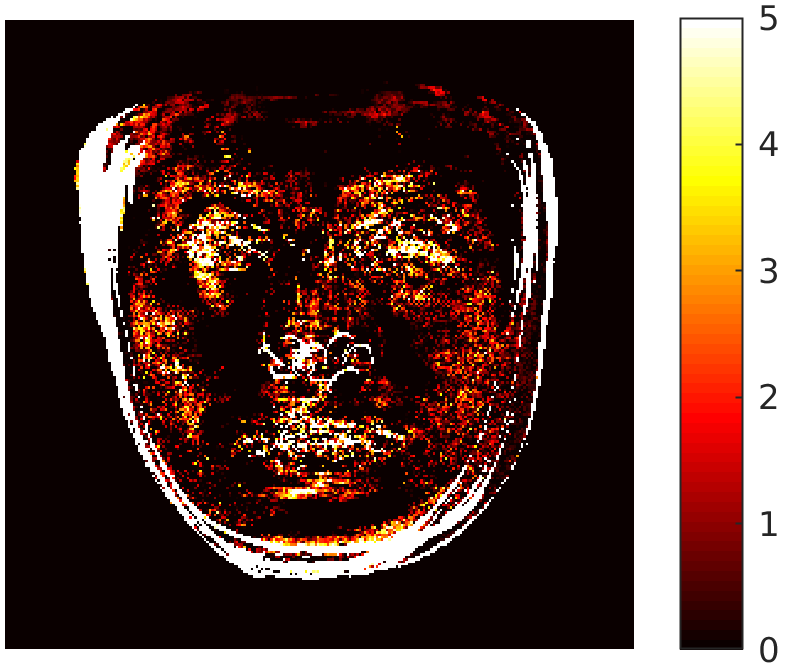}
\end{minipage}
\begin{minipage}[c]{0.2\linewidth}
\centering
\includegraphics[width=\linewidth]{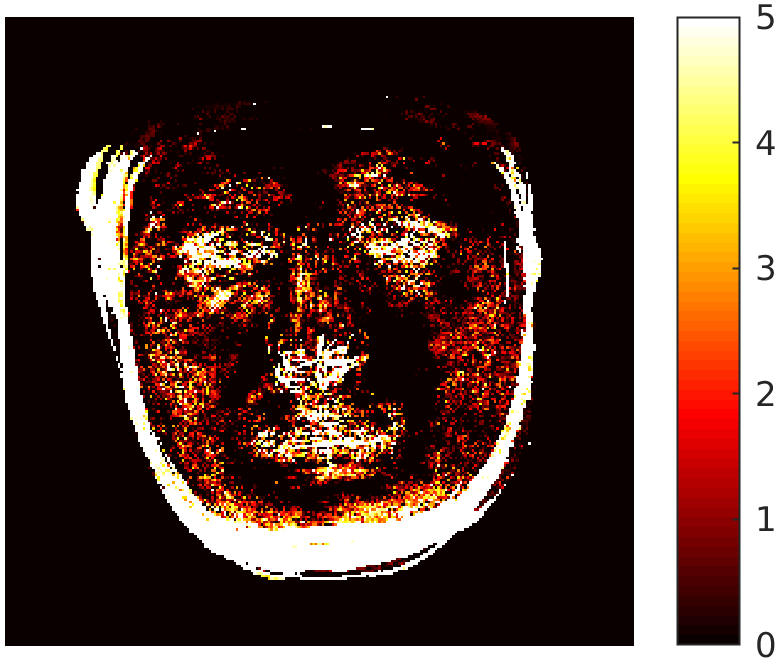}
\end{minipage}
\begin{minipage}[c]{0.2\linewidth}
\centering
\includegraphics[width=\linewidth]{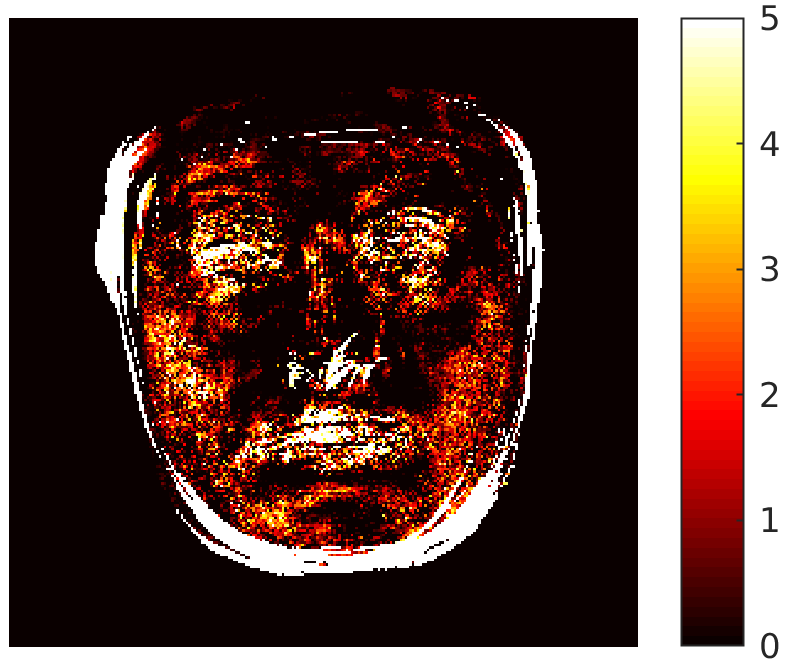}
\end{minipage}
\begin{minipage}[c]{0.2\linewidth}
\centering
\includegraphics[width=\linewidth]{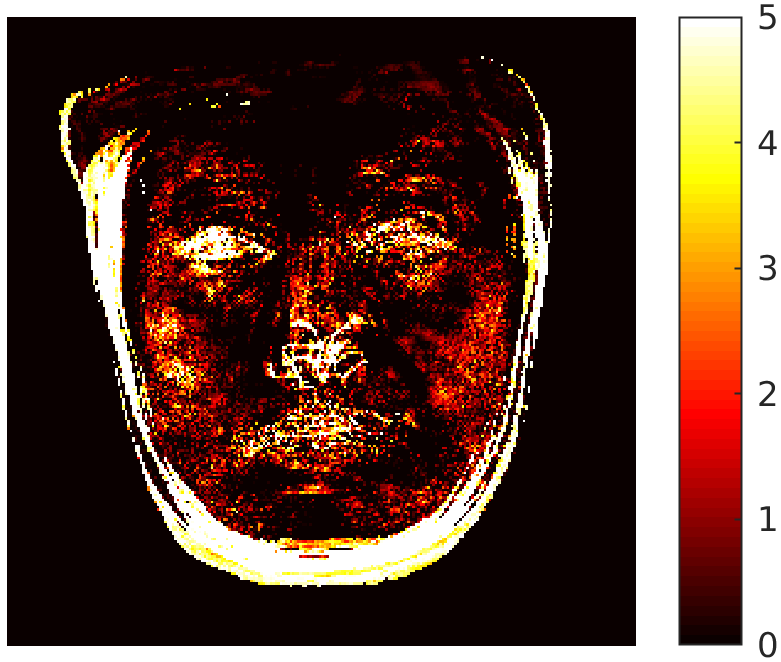}
\end{minipage}
\vspace{-2mm}
\caption{\small\textbf{Surface Normal Improvement}. 
We visualize our surface normal improvement over the best baseline (DFNRMVS~\cite{DFNRMVS}) for test images in $4$ lightings with hard shadows. 
The first row shows a reference image for each lighting. 
Notice the large improvements along and near the nose bridge and the face boundary, which cast hard shadows. 
This shows the contribution of our differentiable hard shadow modeling in improving the geometry.  
\vspace{-6mm}}\label{fig:GeometryErrorVisualization}
\end{center}
\vspace{-2mm}
\end{figure}

As shown in Tab.~\ref{tab:SurfaceNormalEvaluation}, across all test images and lightings, our model achieves the lowest average angular error in surface normal estimation. 
Improving over DFNRMVS shows that our model is not upper bounded by the quality of our shape supervision.
Our end-to-end training incorporating differentiable shadow modeling can yield further improvements to the geometry. 
Improving over SfSNet also highlights the contribution of our shadow modeling, as SfSNet uses a diffuse SH lighting model and thus has no incentive to improve the geometry in regions producing hard shadows. 
Fig.~\ref{fig:GeometryAbsoluteDepth} further demonstrates that the largest improvements are achieved for lightings with significant hard shadows. 
Fig.~\ref{fig:GeometryErrorVisualization} visualizes that our model improves the geometry of the nose and especially the nose bridge significantly, which is where hard shadows are cast from. 
It also improves near the boundary of the face, which also tends to produce hard shadows. 
This demonstrates that our differentiable hard shadow modeling improves the geometry estimation, especially in regions that cast hard shadows.

\begin{table}[t!]
\tiny
\begin{center}
\scalebox{1.25}{
\begin{tabular}{ c | c }
\hline
Method & Surface Normal Angular Error (Degrees) \\ 
\hline
\hline
SfSNet \cite{SfSNet} & $14.2796\!\pm\!2.1442$ \\
DFNRMVS \cite{DFNRMVS} & $12.4505\!\pm\!2.3939$ \\
Proposed & $\mathbf{11.0672\!\pm\!1.9489}$ \\
\hline
\end{tabular}}
\end{center}
\vspace{-6mm}
\caption{\small
\textbf{Surface Normal Errors on Multi-PIE (mean$\pm$ standard deviation)}. We compare with SfSNet and DFNRMVS. 
Our model produces more accurate surface normals than SfSNet, which assumes a diffuse SH lighting model, and DFNRMVS, our shape supervision, which shows the ability of our differentiable hard shadow modeling in improving the geometry. 
}\label{tab:SurfaceNormalEvaluation}
\vspace{-2mm}
\end{table}

\vspace{-2mm}
\Section{Conclusion}

We have proposed a novel face relighting method that produces geometrically consistent hard shadows. Unlike prior work, our approach is the first to directly synthesize cast shadows from the geometry, which improves the shadow's shape and boundary. We have shown on the Multi-PIE and FFHQ datasets that our method achieves state-of-the-art face relighting performance quantitatively and qualitatively under directional lighting. We have also shown that our differentiable hard shadow modeling improves the geometry, especially around the nose, compared to prior work that assumes diffuse shading. We hope that our work will motivate future physics-driven relighting methods, and provide insights for handling hard shadows. 

\Paragraph{Limitations}
Since we use SfSNet's \cite{SfSNet} imperfect estimated lighting as supervision, our model accumulates more error during lighting transfer. The RGB albedo from SfSNet also does not generalize well to our training data, limiting the quality of our estimated albedo. Training on a dataset where we know the groundtruth lightings and could compute the albedo from photometric stereo similar to \cite{PhysicsGuidedRelighting} would improve our model's performance. In addition, although CelebA-HQ \cite{CelebA-HQ} contains some images under directional lighting, it primarily contains images under diffuse lighting. Our model would benefit from a publicly available in-the-wild dataset with primarily directional lights. 

\Paragraph{Broader Impact} 
Creating deepfakes or affecting surveillance by adding shadows are major concerns. We acknowledge these risks but argue that our model only adds self shadows, which are generally limited in size and primarily around the nose. The user cannot freely manipulate the image or add shadows to any location they desire, which limits malicious use cases. 
Moreover, our method can synthesize images with self shadows for training, which can improve the robustness of face methods to self-shadowed images. 

{\small
\bibliographystyle{ieee_fullname}
\bibliography{egbib}
}

\clearpage
\setcounter{equation}{0}
\setcounter{figure}{0}
\setcounter{table}{0}
\setcounter{section}{0}
\twocolumn[\centering \section*{\Large \textbf{Face Relighting with Geometrically Consistent Shadows \\ (Supplementary Materials)\\[1cm]}}] 

\begin{table}[t]
\setlength\tabcolsep{4pt}{ 
\begin{center}
\begin{tabular}{ c | c | c | c }
\hline
\small Method & \small MSE & \small DSSIM & \small LPIPS \\
\hline
\hline
\small DPR \cite{DPR} & \small $0.0171$ & \small $0.0796$ & \small $0.1286$ \\
\small Proposed & \small $\mathbf{0.0080}$ & \small $\mathbf{0.0562}$ & \small $\mathbf{0.1268}$ \\
\hline
\end{tabular}
\end{center}
\vspace{-4mm}
}\caption{\small
\textbf{Quantitative Evaluation for General (Diffuse) Relighting} We outperform DPR quantitatively across all metrics in diffuse relighting on the Multi-PIE dataset. 
}\label{tab:GeneralLightingsQuantitative}
\vspace{-1mm}
\end{table}

\begin{figure}
\includegraphics[width=0.99\linewidth]{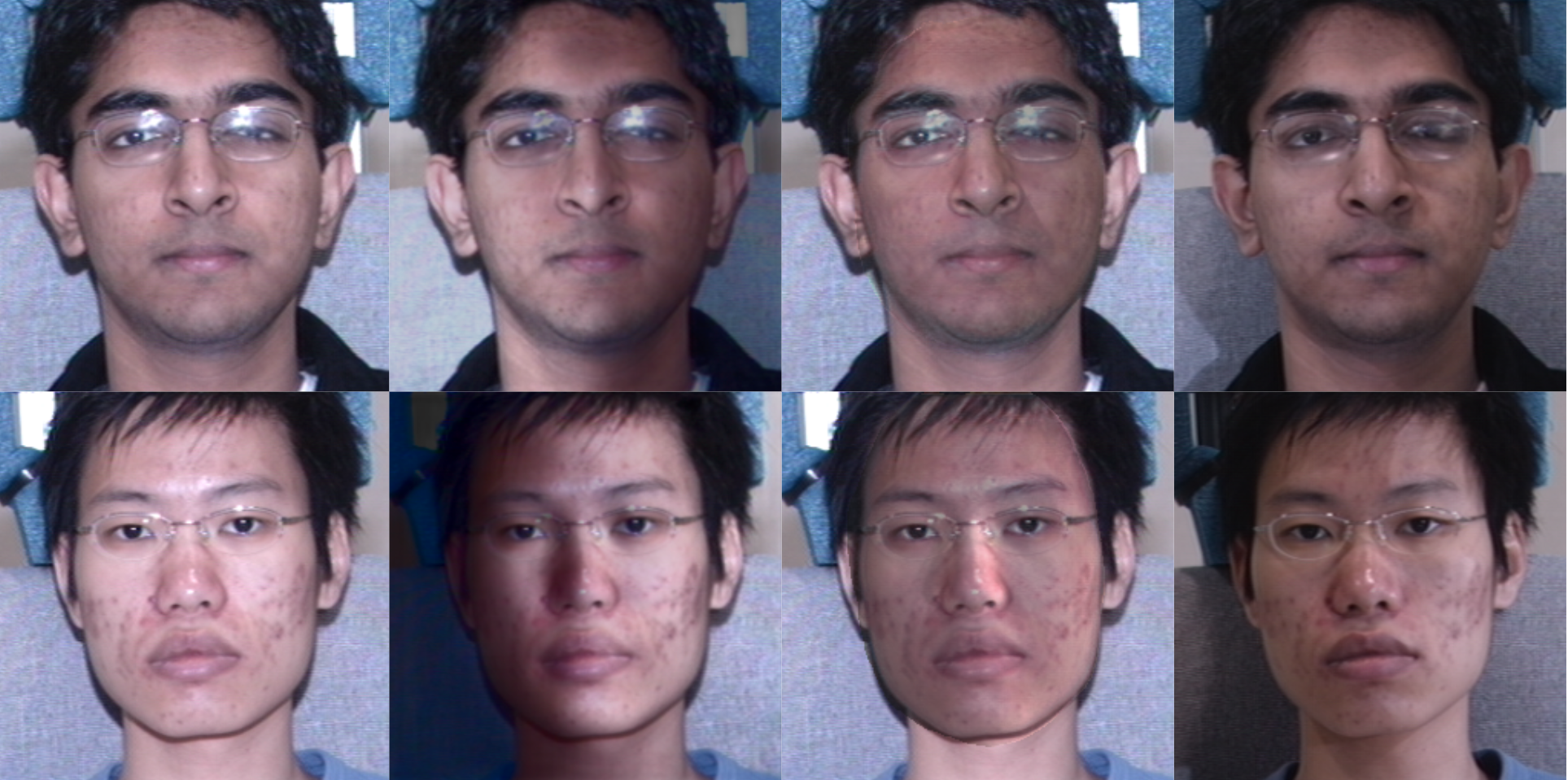} \\
\centering
\small Input \quad\quad\quad\:\: DPR \cite{DPR} \quad\quad\:\: Proposed \quad\quad\: Target
\vspace{-1mm}
\caption{\small \textbf{General (Diffuse) Relighting}. We outperform DPR qualitatively in diffuse relighting on the Multi-PIE dataset, where each input image is relit by averaging the predictions of $3$ randomly selected target lightings. The groundtruth is the average of the $3$ groundtruth Multi-PIE images. 
}\label{fig:GenericLightingQualitative}
\vspace{-4mm}
\end{figure}

\section{Diffuse Relighting Evaluation}
To compare with DPR \cite{DPR} on more general, diffuse lightings, we follow their protocol and generate diffuse lighting groundtruth by averaging $3$ random directional lighting images per Multi-PIE \cite{Multi-PIE} subject. For both DPR and our method, we feed each of the $3$ target lightings separately and average the predictions to generate the final relit image. We outperform DPR in general relighting both quantitatively and qualitatively (Tab.~\ref{tab:GeneralLightingsQuantitative} and Fig.~\ref{fig:GenericLightingQualitative}). 
\section{Geometric Consistency Comparison with Nestmeyer \textit{et al.} \cite{PhysicsGuidedRelighting}}
\begin{figure}
\includegraphics[width=0.9\linewidth]{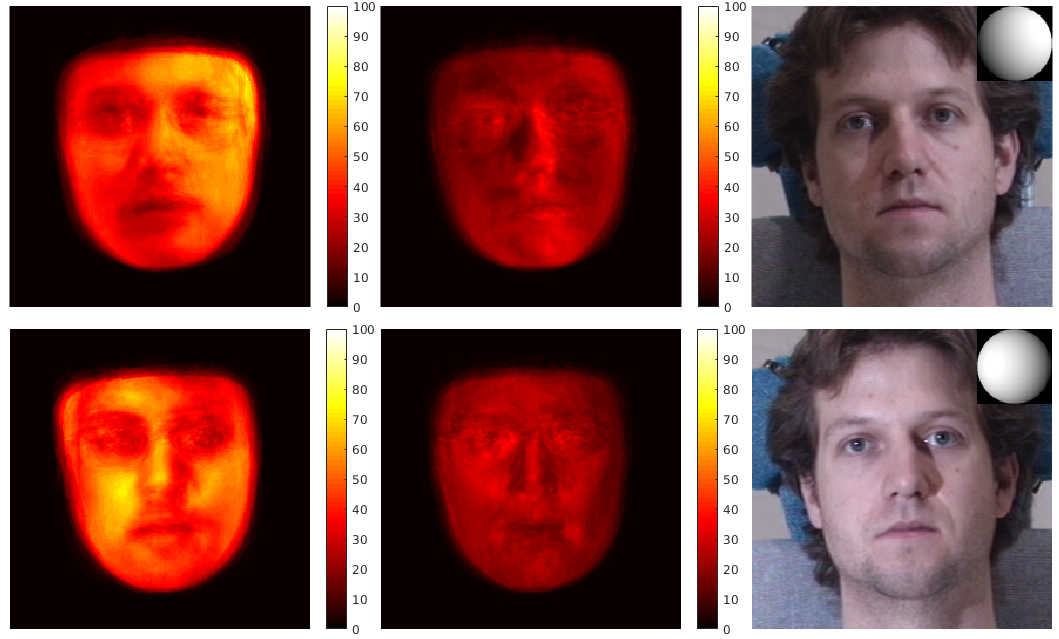} \\
\centering
\small \quad Nestmeyer \cite{PhysicsGuidedRelighting} \quad\quad\quad Proposed \quad\quad\:\: Target Lighting
\vspace{-1mm}
\caption{\small \textbf{Error Maps}. We visualize the average $L_{1}$ error for each Multi-PIE lighting's test subjects. Our method has significantly lower error around the hard shadow regions (nose and cheek) compared to Nestmeyer \textit{et al.} \cite{PhysicsGuidedRelighting}, which demonstrates that our method produces more geometrically consistent hard shadows.  
\vspace{-4mm}
}\label{fig:NestmeyerError}
\end{figure}

To compare with the SoTA relighting method Hou \textit{et al.} \cite{ShadowMaskFaceRelighting}, we used the average $L_{1}$ error for each Multi-PIE lighting's test subjects to verify that our model was improving primarily around the hard shadow region in Fig.~$6$ of the main paper. We show the same error map to compare with Nestmeyer \textit{et al.} \cite{PhysicsGuidedRelighting} in Fig.~\ref{fig:NestmeyerError}. Our method has particularly low error in the hard shadow region (nose and cheek), whereas Nestmeyer \textit{et al.} has high error in and around the shadow, especially for the first row's lighting. Our method thus produces more geometrically consistent hard shadows. 
\vspace{-2mm}
\section{Albedo Comparison}
\begin{figure}
\includegraphics[width=0.9\linewidth]{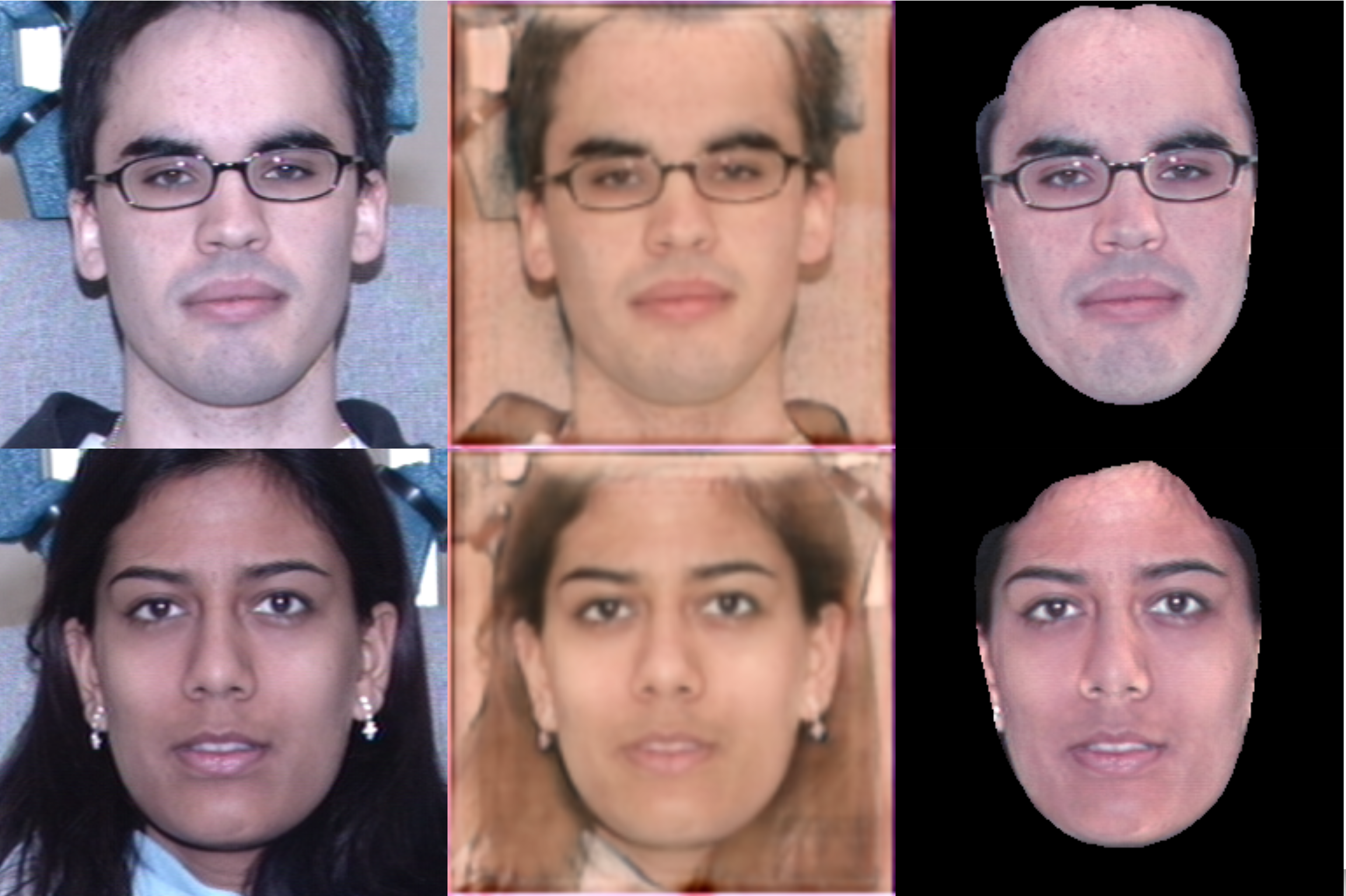} \\
\centering
\small Input Image \quad\quad\quad\: SfSNet \cite{SfSNet} \quad\quad\quad Proposed
\vspace{-1mm}
\caption{\small \textbf{Albedo Comparison}. Our method is able to produce high quality albedo despite the imperfect supervision from SfSNet \cite{SfSNet} by keeping the albedo loss $\mathcal{L}_{albedo}$ in grayscale, which gives our model more freedom in the RGB space. 
\vspace{-4mm}
}\label{fig:IntrinsicComponents}
\end{figure}

Our albedo supervision from SfSNet \cite{SfSNet} is far from perfect, as shown in Fig.~\ref{fig:IntrinsicComponents}, which is why we define the albedo loss in grayscale and not RGB. We adopt this supervision primarily because albedo supervision has limited options for single image in-the-wild datasets besides PCA, which often does not preserve facial details well. However, our model's estimated albedo clearly improves over SfSNet. \\
\vspace{-3mm}
\section{Comprehensive FFHQ Relighting Results}
\begin{figure*}
\vspace{-0.4in}
\begin{center}
\begin{minipage}[c]{0.02\linewidth}
a)
\end{minipage}
\begin{minipage}[c]{0.135\linewidth}
\centering
\includegraphics[width=\linewidth]{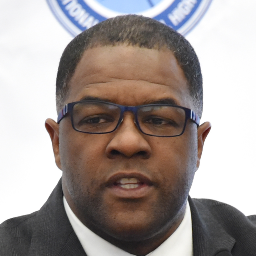}
\end{minipage}
\begin{minipage}[c]{0.135\linewidth}
\centering
\includegraphics[width=\linewidth]{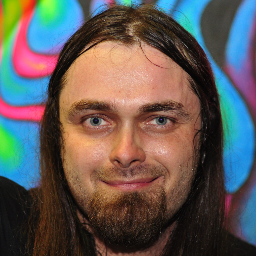}
\end{minipage}
\begin{minipage}[c]{0.135\linewidth}
\centering
\includegraphics[width=\linewidth]{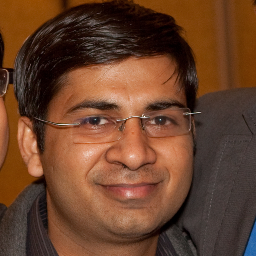}
\end{minipage}
\begin{minipage}[c]{0.135\linewidth}
\centering
\includegraphics[width=\linewidth]{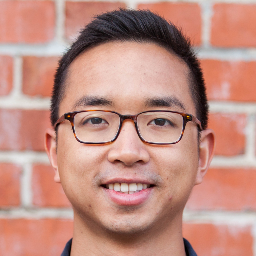}
\end{minipage}
\begin{minipage}[c]{0.135\linewidth}
\includegraphics[width=\linewidth]{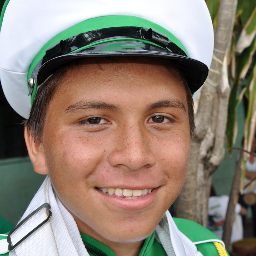}
\end{minipage}
\begin{minipage}[c]{0.135\linewidth}
\centering
\includegraphics[width=\linewidth]{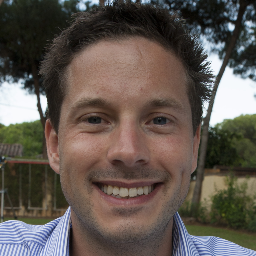}
\end{minipage}
\begin{minipage}[c]{0.135\linewidth}
\centering
\includegraphics[width=\linewidth]{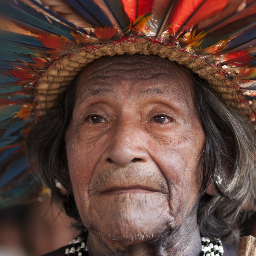}
\end{minipage}

\vspace{0.5mm}
\begin{minipage}[c]{0.02\linewidth}
b)
\end{minipage}
\begin{minipage}[c]{0.135\linewidth}
\centering
\includegraphics[width=\linewidth]{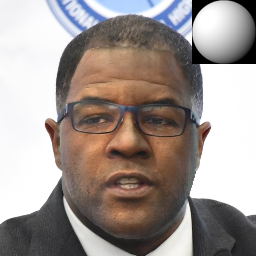}
\end{minipage}
\begin{minipage}[c]{0.135\linewidth}
\centering
\includegraphics[width=\linewidth]{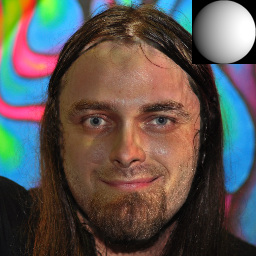}
\end{minipage}
\begin{minipage}[c]{0.135\linewidth}
\centering
\includegraphics[width=\linewidth]{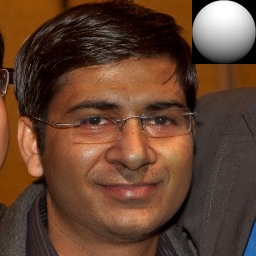}
\end{minipage}
\begin{minipage}[c]{0.135\linewidth}
\centering
\includegraphics[width=\linewidth]{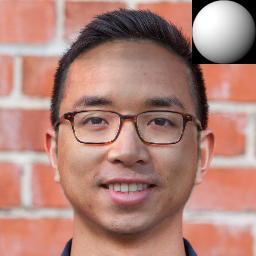}
\end{minipage}
\begin{minipage}[c]{0.135\linewidth}
\centering
\includegraphics[width=\linewidth]{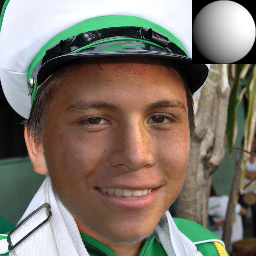}
\end{minipage}
\begin{minipage}[c]{0.135\linewidth}
\centering
\includegraphics[width=\linewidth]{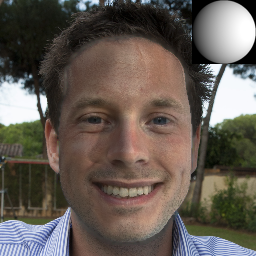}
\end{minipage}
\begin{minipage}[c]{0.135\linewidth}
\centering
\includegraphics[width=\linewidth]{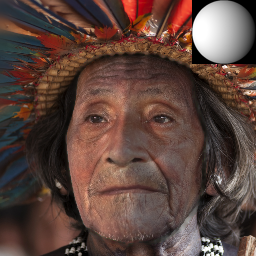}
\end{minipage}

\vspace{0.5mm}
\begin{minipage}[c]{0.02\linewidth}
c)
\end{minipage}
\begin{minipage}[c]{0.135\linewidth}
\centering
\includegraphics[width=\linewidth]{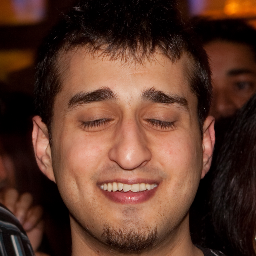}
\end{minipage}
\begin{minipage}[c]{0.135\linewidth}
\centering
\includegraphics[width=\linewidth]{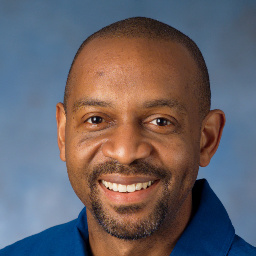}
\end{minipage}
\begin{minipage}[c]{0.135\linewidth}
\centering
\includegraphics[width=\linewidth]{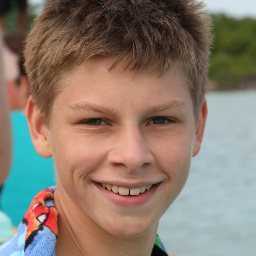}
\end{minipage}
\begin{minipage}[c]{0.135\linewidth}
\centering
\includegraphics[width=\linewidth]{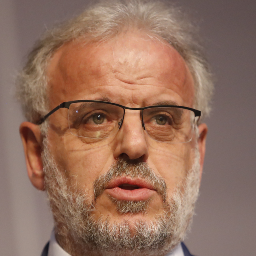}
\end{minipage}
\begin{minipage}[c]{0.135\linewidth}
\centering
\includegraphics[width=\linewidth]{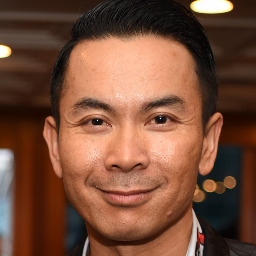}
\end{minipage}
\begin{minipage}[c]{0.135\linewidth}
\centering
\includegraphics[width=\linewidth]{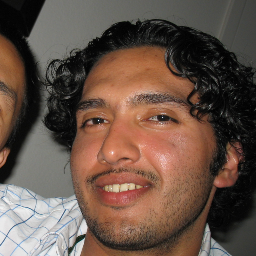}
\end{minipage}
\begin{minipage}[c]{0.135\linewidth}
\centering
\includegraphics[width=\linewidth]{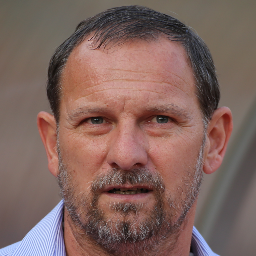}
\end{minipage}

\vspace{0.5mm}
\begin{minipage}[c]{0.02\linewidth}
d)
\end{minipage}
\begin{minipage}[c]{0.135\linewidth}
\centering
\includegraphics[width=\linewidth]{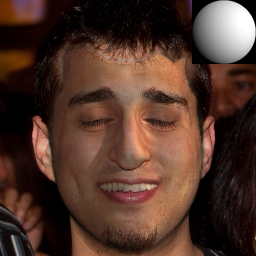}
\end{minipage}
\begin{minipage}[c]{0.135\linewidth}
\centering
\includegraphics[width=\linewidth]{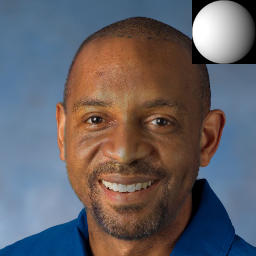}
\end{minipage}
\begin{minipage}[c]{0.135\linewidth}
\centering
\includegraphics[width=\linewidth]{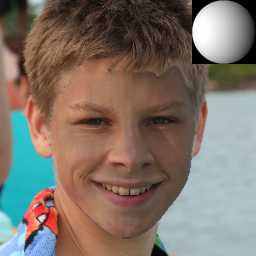}
\end{minipage}
\begin{minipage}[c]{0.135\linewidth}
\centering
\includegraphics[width=\linewidth]{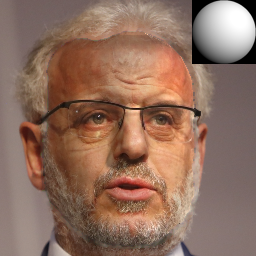}
\end{minipage}
\begin{minipage}[c]{0.135\linewidth}
\centering
\includegraphics[width=\linewidth]{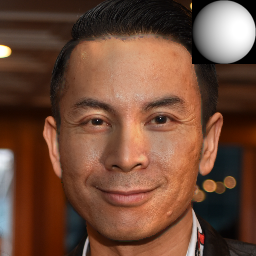}
\end{minipage}
\begin{minipage}[c]{0.135\linewidth}
\centering
\includegraphics[width=\linewidth]{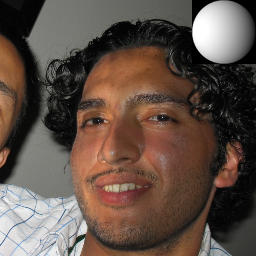}
\end{minipage}
\begin{minipage}[c]{0.135\linewidth}
\centering
\includegraphics[width=\linewidth]{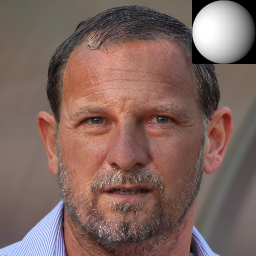}
\end{minipage}

\vspace{0.5mm}
\begin{minipage}[c]{0.02\linewidth}
e)
\end{minipage}
\begin{minipage}[c]{0.135\linewidth}
\centering
\includegraphics[width=\linewidth]{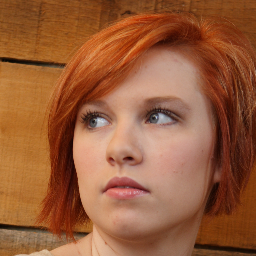}
\end{minipage}
\begin{minipage}[c]{0.135\linewidth}
\centering
\includegraphics[width=\linewidth]{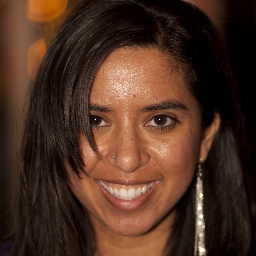}
\end{minipage}
\begin{minipage}[c]{0.135\linewidth}
\centering
\includegraphics[width=\linewidth]{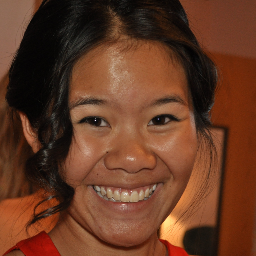}
\end{minipage}
\begin{minipage}[c]{0.135\linewidth}
\centering
\includegraphics[width=\linewidth]{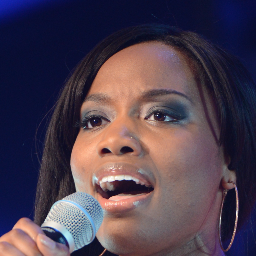}
\end{minipage}
\begin{minipage}[c]{0.135\linewidth}
\centering
\includegraphics[width=\linewidth]{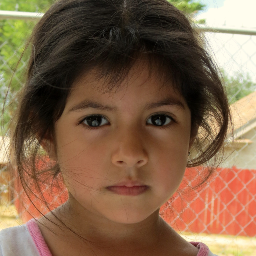}
\end{minipage}
\begin{minipage}[c]{0.135\linewidth}
\centering
\includegraphics[width=\linewidth]{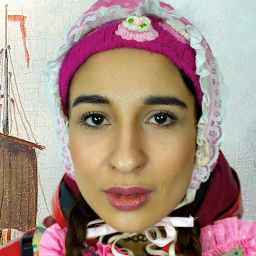}
\end{minipage}
\begin{minipage}[c]{0.135\linewidth}
\centering
\includegraphics[width=\linewidth]{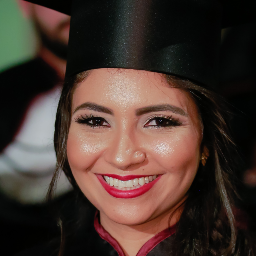}
\end{minipage}

\vspace{0.5mm}
\begin{minipage}[c]{0.02\linewidth}
f)
\end{minipage}
\begin{minipage}[c]{0.135\linewidth}
\centering
\includegraphics[width=\linewidth]{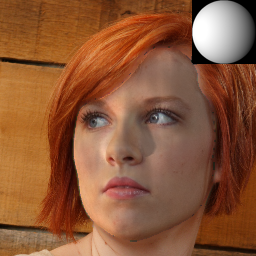}
\end{minipage}
\begin{minipage}[c]{0.135\linewidth}
\centering
\includegraphics[width=\linewidth]{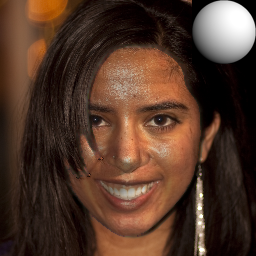}
\end{minipage}
\begin{minipage}[c]{0.135\linewidth}
\centering
\includegraphics[width=\linewidth]{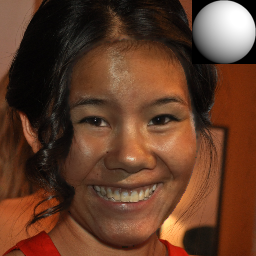}
\end{minipage}
\begin{minipage}[c]{0.135\linewidth}
\centering
\includegraphics[width=\linewidth]{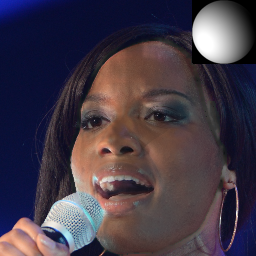}
\end{minipage}
\begin{minipage}[c]{0.135\linewidth}
\centering
\includegraphics[width=\linewidth]{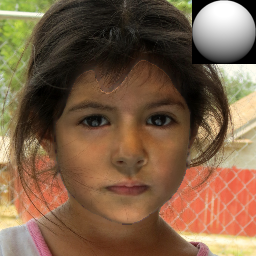}
\end{minipage}
\begin{minipage}[c]{0.135\linewidth}
\centering
\includegraphics[width=\linewidth]{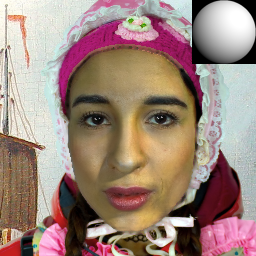}
\end{minipage}
\begin{minipage}[c]{0.135\linewidth}
\centering
\includegraphics[width=\linewidth]{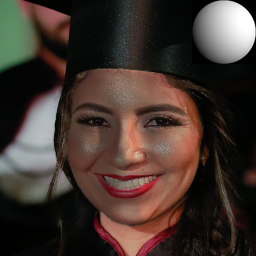}
\end{minipage}

\vspace{0.5mm}
\begin{minipage}[c]{0.02\linewidth}
g)
\end{minipage}
\begin{minipage}[c]{0.135\linewidth}
\centering
\includegraphics[width=\linewidth]{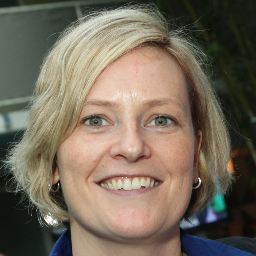}
\end{minipage}
\begin{minipage}[c]{0.135\linewidth}
\centering
\includegraphics[width=\linewidth]{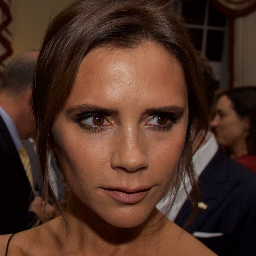}
\end{minipage}
\begin{minipage}[c]{0.135\linewidth}
\centering
\includegraphics[width=\linewidth]{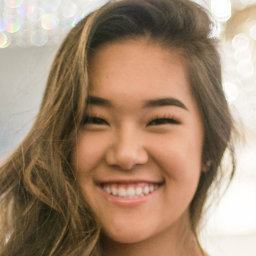}
\end{minipage}
\begin{minipage}[c]{0.135\linewidth}
\centering
\includegraphics[width=\linewidth]{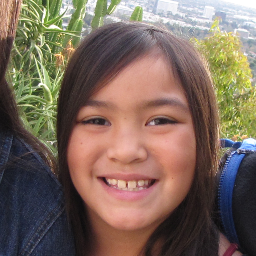}
\end{minipage}
\begin{minipage}[c]{0.135\linewidth}
\centering
\includegraphics[width=\linewidth]{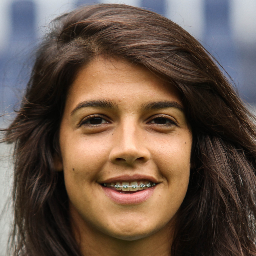}
\end{minipage}
\begin{minipage}[c]{0.135\linewidth}
\centering
\includegraphics[width=\linewidth]{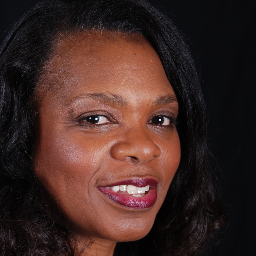}
\end{minipage}
\begin{minipage}[c]{0.135\linewidth}
\centering
\includegraphics[width=\linewidth]{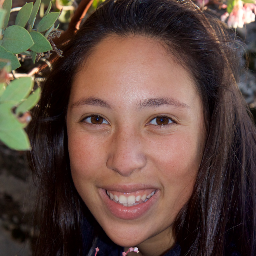}
\end{minipage}

\vspace{0.5mm}
\begin{minipage}[c]{0.02\linewidth}
h)
\end{minipage}
\begin{minipage}[c]{0.135\linewidth}
\centering
\includegraphics[width=\linewidth]{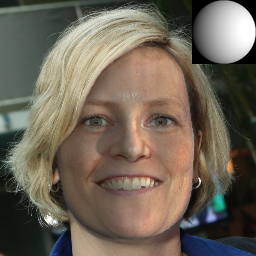}
\end{minipage}
\begin{minipage}[c]{0.135\linewidth}
\centering
\includegraphics[width=\linewidth]{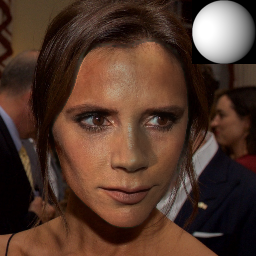}
\end{minipage}
\begin{minipage}[c]{0.135\linewidth}
\centering
\includegraphics[width=\linewidth]{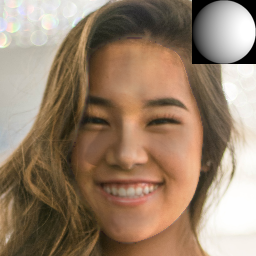}
\end{minipage}
\begin{minipage}[c]{0.135\linewidth}
\centering
\includegraphics[width=\linewidth]{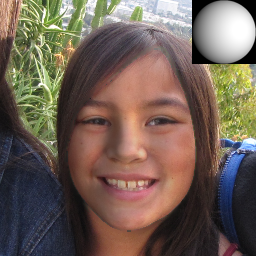}
\end{minipage}
\begin{minipage}[c]{0.135\linewidth}
\centering
\includegraphics[width=\linewidth]{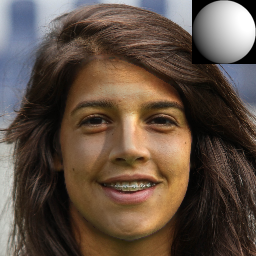}
\end{minipage}
\begin{minipage}[c]{0.135\linewidth}
\centering
\includegraphics[width=\linewidth]{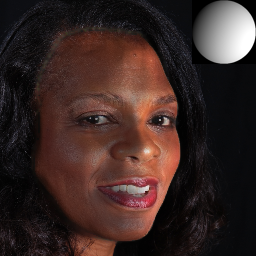}
\end{minipage}
\begin{minipage}[c]{0.135\linewidth}
\centering
\includegraphics[width=\linewidth]{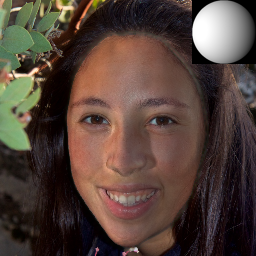}
\end{minipage}

\vspace{-1mm}
\caption{\small\textbf{Comprehensive and Diverse Relighting Performance on FFHQ}. Every two rows (\textit{e.g.} c, d) shows the input image in the first row and our relighting results in the second row. We demonstrate our relighting performance on a wide variety of racial groups, genders, ages, expressions, and poses and also include subjects with facial hair and glasses. We find that our model is able to generalize to a wide range of subjects across many different lightings. Best viewed if enlarged.}\label{fig:FFHQSupp}
\end{center}
\end{figure*}

We strongly believe in diversity and the representation of all groups in the computer vision community. We therefore show a wide variety of relighting results with diversity and inclusion in mind. Our results cover as many racial groups as possible, as well as other factors such as different ages, genders, poses, expressions, subjects with facial hair, and the presence of glasses (See Fig.~\ref{fig:FFHQSupp}). We also increased the lighting diversity to demonstrate that our model can handle many different desired illuminations. 

\section{FFHQ Relighting Video}
We include a video with $4$ FFHQ \cite{FFHQ} subjects where we rotate the light around the face, move the light horizontally, and move the light vertically. From left to right, we visualize the target lighting, the relighting results of Hou \textit{et al.} \cite{ShadowMaskFaceRelighting}, and our proposed method's relighting results. Our video demonstrates our high relighting quality as well as the geometric consistency of our shadows across many lightings. Compared to \cite{ShadowMaskFaceRelighting}, it is clear that the shape of our shadows is superior, especially when comparing the first subject. We also modify the tone of the image significantly less, while \cite{ShadowMaskFaceRelighting} seems to frequently produce overly dark shadows. The video can be viewed \href{https://www.youtube.com/watch?v=fbP2i5ywZvw}{here}. 
\section{Licenses for Face Related Datasets}
Although we don't collect any face data ourselves in this work, we do make use of existing face datasets, including Multi-PIE \cite{Multi-PIE}, FFHQ \cite{FFHQ}, and CelebA-HQ \cite{CelebA-HQ}. The Multi-PIE database was collected at Carnegie Mellon University, where all subjects agreed that their data would be used for research purposes. We only use the database internally for our work and primarily for evaluation. FFHQ consists of images published on Flickr, which are all under multiple licenses that allow free use, adaptation, and redistribution for noncommercial purposes. The creators also provide a way to remove an individual's photo from the dataset if they so desire. CelebA-HQ consists entirely of images collected from the internet. Although there is no associated IRB approval, the authors assert in the dataset agreement that the dataset is only to be used for noncommercial research purposes, which we strictly adhere to. Users must also agree not to sell, reproduce, or exploit any of the data and can only make copies of the data within their own organization, which we also adhere to. 
\newpage
\end{document}